\pdfoutput=1
\pdfoutput=1
\pdfoutput=1
\pdfoutput=1
\pdfoutput=1
\documentclass[twocolumn,10pt]{IEEEtran}
\normalsize

\usepackage{enumerate}
\usepackage{color}

\usepackage{cuted}
\usepackage{cite}

\usepackage{float}
\newfloat{figtab}{htb}{fgtb}
\makeatletter
\newcommand\figcaption{\def\@captype{figure}\caption}
\newcommand\tabcaption{\def\@captype{table}\caption}
\makeatother

\usepackage{graphicx}
\usepackage{booktabs}
\usepackage{algorithm}
\usepackage{bm}
\ifCLASSINFOpdf
\else
\fi

\usepackage{amsthm}
\usepackage{amssymb,amsfonts}
\usepackage{amsfonts,balance}

\usepackage[cmex10]{amsmath}
\usepackage{lipsum}

\newtheorem{theorem}{Theorem}

\usepackage{algorithmic}

\usepackage{array}
\usepackage{multirow,makecell}
\usepackage[caption=false,font=footnotesize]{subfig}

\usepackage{verbatim}

\begin{document}
%
\title{Adaptive Model Pruning and Personalization for Federated Learning over Wireless Networks}
%
%
\author{Xiaonan~Liu,~\IEEEmembership{Member,~IEEE,} Tharmalingam Ratnarajah,~\IEEEmembership{Senior~Member,~IEEE}, Mathini Sellathurai,~\IEEEmembership{Fellow,~IEEE}, and Yonina C. Eldar,~\IEEEmembership{Fellow,~IEEE}\thanks{X. Liu and T. Ratnarajah are with the Institute for Digital Communications, The University of Edinburgh, U.K. (e-mail: \{xliu8, t.ratnarajah\}@ed.ac.uk). 
}\thanks{M. Sellathurai is with the Department of Engineering
and Physical Science (EPS), Heriot-Watt University, Edinburgh,
U.K. (e-mail: m.sellathurai@hw.ac.uk)}\thanks{Yonina C. Eldar is with the Faculty of Mathematics and Computer
Science, Weizmann Institute of Science, Rehovot 7610001, Israel (e-mail:
yonina.eldar@weizmann.ac.il)}}
\maketitle

\begin{abstract}
Federated learning (FL) enables distributed learning across edge devices while protecting data privacy. However, the learning accuracy decreases due to the heterogeneity of devices' data, and the computation and communication latency increase when updating large-scale learning models on devices with limited computational capability and wireless resources. We consider a FL framework with partial model pruning and personalization to overcome these challenges. This framework splits the learning model into a global part with model pruning shared with all devices to learn data representations and a personalized part to be fine-tuned for a specific device, which adapts the model size during FL to reduce both computation and communication latency and increases the learning accuracy for devices with non-independent and identically distributed data. The computation and communication latency and convergence of the proposed FL framework are mathematically analyzed. To maximize the convergence rate and guarantee learning accuracy, Karush–Kuhn–Tucker (KKT) conditions are deployed to jointly optimize the pruning ratio and bandwidth allocation. Finally, experimental results demonstrate that the proposed FL framework achieves a remarkable reduction of approximately $50\%$ computation and communication latency compared with FL with partial model personalization.
\end{abstract}



\begin{IEEEkeywords}
Adaptive partial model pruning and personalization, communication and computation latency, federated learning, wireless networks.
\end{IEEEkeywords}

%
\IEEEpeerreviewmaketitle


\section{Introduction}
Federated learning (FL) is a promising approach for protecting data privacy and facilitating distributed learning across diverse domains, such as healthcare, finance, and mobile devices \cite{9415623,9460016}, where multiple edge devices share only the model/gradients with an edge server to collaboratively train a learning model \cite{B_McMahan}. Although FL preserves privacy, it still has significant challenges in computation and communication \cite{xiaonan_survey,9851495}. Data heterogeneity in different devices may lead to unstable training of FL and further result in poor generalization ability of the global model and low learning accuracy. Furthermore, devices with limited computational capabilities lead to high computation latency. On the communication side, with an increasing number of edge devices, the limited communication resources, e.g., bandwidth and energy, may impede devices from effectively contributing to model aggregation, which results in high transmission latency and has a negative effect on the convergence rate and learning accuracy \cite{9084352}.

To address data heterogeneity, personalized FL (PFL) and federated feature matching (FedFM) were considered in \cite{jiahaotan,jiahaoliu,10089406} and \cite{10286439}, respectively. In \cite{jiahaotan}, PFL based on model similarity was proposed to leverage a classifier-based similarity for personalized model aggregation without incurring extra communication overhead. In \cite{jiahaoliu}, a new PFL algorithm called FL with dynamic weight adjustment was proposed to leverage the edge server to calculate personalized aggregation weights based on collected models from devices, which captures similarities between devices and improves PFL learning accuracy by encouraging collaboration among devices with similar data distribution. In \cite{10089406}, hierarchical PFL (HPFL) was considered in massive mobile edge computing (MEC) networks. HPFL combined the objectives of training loss minimization and round latency minimization while jointly determining the optimal bandwidth allocation and the edge server scheduling policy in a hierarchical learning framework. In \cite{10286439}, FedFM was proposed to guide each device's features to match shared category-wise anchors (landmarks in feature space), which mitigated the negative effects of data heterogeneity in FL by aligning each device’s feature space. However, PFL and FedFM in \cite{jiahaotan,jiahaoliu,10089406,10286439} still need to transmit the whole learning model to the server for similarity/anchors calculation and model aggregation.

To deal with the high-latency issue in PFL, FL with partial model personalization is introduced in \cite{pmlr-v162-pillutla22a,Mishchenko}. This FL framework splits the learning model into a global part, shared with all devices to learn data representations, and a personalized part, fine-tuned for a specific device based on the heterogeneity of the local dataset. In each global communication round, the edge server broadcasts the current global part to devices. Then, each device performs one or more steps of stochastic gradient descent to update both the global and personalized parts and transmits only the updated global part to the edge server for model aggregation, further reducing the transmission latency. Meanwhile, the updated personalized part is kept locally on the device to initialize another update in the next global communication round. In \cite{pmlr-v162-pillutla22a}, two FL algorithms for training partially personalized models were introduced, where the shared and personal parameters are updated simultaneously or alternately on devices. In \cite{Mishchenko}, the authors proved that through optimizing the split point of the learning model, it is possible to find global and personalized parameters that allow each device to fit its local dataset perfectly and break the curse of data heterogeneity in several settings. However, the communication and computation overheads of partial model personalization in \cite{pmlr-v162-pillutla22a} and \cite{Mishchenko} over wireless networks are not explored.

To improve the learning efficiency, federated pruning was proposed in \cite{9598845,9762360,xiaonan1}, where the model size is adapted during the training/inference phase or both of these two phases to decrease computation and communication overhead. FL pruning offers benefits in model size reduction without sacrificing the learning accuracy of the original model \cite{Horvath}. In \cite{9598845}, model pruning was adopted before local gradient calculation to reduce the local model computation and gradient communication latency in FL over wireless networks. By removing the edge devices with low computation capability and bad channel conditions, device selection was also considered to save the communication overhead and reduce the model aggregation error caused by model pruning. In \cite{9762360}, a low-complexity adaptive pruning method called PruneFL was proposed for resource-limited devices, which determined a desired model size that achieves similar prediction accuracy as the original model but with less training time. In \cite{xiaonan1}, hierarchical FL with model pruning over wireless networks was introduced to reduce the neural network size, decreasing computation and communication latency while guaranteeing a similar learning accuracy as the original model. Nevertheless, the proposed FL pruning methods in \cite{9598845,9762360,xiaonan1} are unsuitable for devices with non-independent and identically distributed (non-IID) data. Therefore, designing an FL framework with high communication and computation efficiency and adaptability to non-IID datasets is essential.

In addition, unlike FL with personalization or pruning, proper resource allocation and device scheduling also enable computation and communication-efficient FL \cite{yang2020energy, 9292468, 9210812, RA1, Gafni}. Specifically, in \cite{yang2020energy}, by optimizing time allocation, bandwidth allocation, power control, and computation frequency, an iterative algorithm was proposed to minimize the total energy consumption for local computation and wireless transmission of FL. In \cite{9292468}, a probabilistic device selection scheme was proposed to minimize the FL convergence time and the FL training loss in a joint learning, wireless resource allocation, and user selection problem. In \cite{9210812}, the Hungarian algorithm was used to find an optimal device selection and resource block allocation to minimize the FL loss function. In \cite{RA1}, the communication and computation costs were minimized by deploying successive convex approximation and Hungarian algorithms to optimize bandwidth, computation frequency, power allocation, and sub-carrier assignment. In \cite{Gafni}, dubbed Bayesian air aggregation FL (BAAF) was proposed to effectively mitigate noise and fading effects
induced by the channel, and to handle the statistical heterogeneity of devices' data, controlled BAAF was developed to allow for appropriate local updates by the devices. Although the proposed device scheduling and resource allocation approaches in \cite{yang2020energy, 9292468, 9210812, RA1, Gafni} effectively alleviate communication costs, uploading the entire learning model still poses a challenge for devices with poor channel conditions.

Motivated by the aforementioned challenges, we propose a computation and communication-efficient wireless FL framework with partial model pruning and personalization. First, we mathematically model the computation and communication latency of the proposed FL framework. Second, a convergence analysis is presented. Then, Karush–Kuhn–Tucker (KKT) conditions are deployed to derive closed-form solutions for the pruning ratio and bandwidth allocation under latency and bandwidth thresholds, which achieve a high learning accuracy and low computation and communication latency. The main contributions are summarized as follows.

\begin{itemize}
    \item We propose a communication and computation-efficient FL framework with partial model pruning and personalization to adapt to data heterogeneity across different devices and dynamical wireless environments.
    \item We mathematically model the computation and communication latency of the proposed FL framework. We then analyze an upper bound on the $l_2$-norm of gradients of the method. Next, we jointly optimize the pruning ratio of the global part and bandwidth allocation to maximize the convergence rate and guarantee learning accuracy.
    \item Under the latency and bandwidth thresholds, the optimization problem is divided into two sub-problems. Karush–Kuhn–Tucker (KKT) conditions are used to solve these sub-problems and calculate the optimal solutions of the pruning ratio and bandwidth allocation.
    \item Experimental results demonstrate that our proposed FL framework achieves comparable learning accuracy to FL only with model personalization and leads to a reduction of approximately $50\%$ in computation and communication latency.
\end{itemize}

In the rest of this paper, the system model and problem formulation are presented in Section II. Convergence analysis and problem transformation are presented in Section III. Pruning ratio and bandwidth fraction optimization are presented in Section IV. Experimental results and conclusions are given in Section V and Section VI, respectively.

\section{System Model and Problem Formulation}
In this section, we detail the system model for FL with partial model personalization over wireless networks.

\subsection{FL with Partial Model Personalization Architecture}
We consider an edge server with multiple antennas providing wireless service for $\mathcal{K} = \{k = 1,2,...,K\}$ mobile devices. The $k$th edge device has a dataset $\mathcal{D}_{k}$ and $D_k$ is the number of data samples. Without loss of generality, we assume that there is no overlap between datasets from different devices, namely, $\mathcal{D}_n\cap\mathcal{D}_k = \emptyset, (\forall n,k\in\mathcal{K})$. The whole dataset and the total number of data samples are denoted as $\mathcal{D} = \cup\{\mathcal{D}_{k}\}_{k=1}^{K}$ and $D = \sum_{k=1}^{K}D_k$, respectively.

The main objective of traditional FL algorithms, such as federated averaging (FedAvg) \cite{B_McMahan}, is to find an optimal global model $\bm{w}^{*}$ that minimizes the global loss function $F(\bm{w})$:
\begin{equation}
    F(\bm{w}) = \sum_{k=1}^{K}\frac{D_k}{D}F_{k}(\bm{w}),
\end{equation}
where the local loss $F_{k}(\bm{w})$ of the $k$th device is calculated as
\begin{equation}
    F_{k}(\bm{w}) = \frac{1}{D_k}\sum_{i=1}^{D_k}f_{k}(\bm{x}_i,y_i,\bm{w}).
\end{equation}
In (2), $f_{k}(\bm{x}_i,y_i,\bm{w})$ is the loss function, $\bm{x}_i$ is the $i$th input data sample, and $y_i$ is the corresponding labeled output.

The objective function in (1) may not have a global model $\bm{w}$ that fits all devices in practical FL systems, especially when the data distribution in devices is non-IID or heterogeneous, namely, statistical heterogeneity. The local optimal models may drift significantly from each other and lead to a poor generalization of each device because of statistical data heterogeneity. Modern deep learning models usually have a multi-layer architecture. Generally, lower layers (close to the input) are responsible for feature extraction, and upper layers (close to the output) focus on complex pattern recognition. Therefore, based on the application domain, often either the input layer or the output layer of the model can be personalized. We may thus split the model $\bm{w}$ into a global part $\bm{u}$ shared by all devices and a personalized part $\bm{v}_{k}$ specific to the $k$th device, i.e., $\bm{w} = [\bm{u}, \bm{v}_{k}].$ The objective functions in (1) and (2) can then be rewritten as
\begin{equation}
    \min_{\bm{u}, \{\bm{v}_{k}\}_{k=1}^{K}}F(\bm{u}, \bm{v}_{k}) = \sum_{k=1}^{K}\frac{D_k}{D}F_{k}(\bm{u}, \bm{v}_{k}),
\end{equation}
and
\begin{equation}
    F_{k}(\bm{u}, \bm{v}_{k}) = \frac{1}{D_k}\sum_{i=1}^{D_k}f_{k}(\bm{x}_i,y_i,\bm{u}, \bm{v}_{k}).
\end{equation}

\subsection{Learning Process}
In FL with partial model personalization, the personalized parts are only updated in local devices, and the global parts are updated in local devices and aggregated in the edge server. Therefore, the learning process includes global part broadcasting, local model updating, and global part uploading. To quantify the training overhead of the proposed FL framework, we model computation and communication latency. The learning process is detailed as follows:

\subsubsection{Global Part Broadcasting} In the $g$th global communication round, the edge server broadcasts the global part $\bm{u}^{g}$ to devices by downlink transmission. The downlink transmission rate is very high in a practical system due to sufficient channel bandwidth. Consequently, the transmission latency of global part broadcasting is neglected.

\subsubsection{Local Model Updating} 
Local model updating includes personalized and global parts updating. According to \cite{pmlr-v162-pillutla22a}, updating the personalized and global parts alternatively, so-called LocalAlt, guarantees a higher learning accuracy. In LocalAlt, the personalized part $\bm{v}_{k}^{g}$ of the $k$th device is first updated $\tau_{\bm{v}}$ iterations, and the global part remains the same. Then, the global part $\bm{u}_{k}^{g}$ updates $\tau_{\bm{u}}$ iterations using the updated personalized part $\bm{v}_{k}^{g,\tau_{\bm{v}}}$. 

Based on (3) and (4), when the $k$th edge device obtains the global part from the edge server, the objective of the $k$th device is to find
\begin{equation}
    \bm{v}_{k}^{*}(\bm{u}_{k}^{g}) = \arg\min_{\bm{v}_{k}^{g}}F_{k}(\bm{u}_{k}^{g}, \bm{v}_{k}^{g}).
\end{equation}
It is possible that the edge device has limited computation capacity; thus, updating personalized and global parts over the whole dataset is time-consuming. As a result, we consider stochastic gradient descent (SGD) where the $k$th edge device utilizes a subset of its dataset to update the local model. Within the $g$th global communication round, the personalized part updating in the $t$th iteration is written as
\begin{equation}
    \bm{v}_{k}^{g,t+1} = \bm{v}_{k}^{g,t} - \eta_{\bm{v}}\nabla_{\bm{v}} F_{k}(\bm{u}_{k}^{g}, \bm{v}_{k}^{g,t}, \xi_{k}^{g,t}),
\end{equation}
where $\nabla_{\bm{v}} F_{k}(\bm{u}_{k}^{g}, \bm{v}_{k}^{g,t}, \xi_{n}^{g,t})$ and $\eta_{\bm{v}}$ are the gradient and learning rate of the personalized part, respectively, and $\xi_{k}^{g,t}\subseteq\mathcal{D}_{k}$ is the randomly selected mini-batch datasets. Based on the updated personalized part $\bm{v}_{k}^{g,\tau_{\bm{v}}}$, the global part updating in the $t$th iteration is denoted as
\begin{equation}
    \bm{u}_{k}^{g,t+1} = \bm{u}_{k}^{g,t} - \eta_{\bm{u}}\nabla F_{k}(\bm{u}_{k}^{g,t}, \bm{v}_{k}^{g,\tau_{\bm{v}}}, \xi_{k}^{g,t}),
\end{equation}
where $\nabla F_{k}(\bm{u}_{k}^{g,t}, \bm{v}_{k}^{g,\tau_{\bm{v}}}, \xi_{k}^{g,t})$ and $\eta_{\bm{u}}$ are the gradient and learning rate of the global part, respectively.

Assuming that the number of CPU cycles to update one model weight and the allocated CPU frequency of the $k$th device are $C_k$ and $f_{k}\in[f_k^{\min}, f_k^{\max}]$, respectively, the computation latency of personalized and global parts updating is calculated as 
\begin{equation}
    T_{k, g}^{\text{cmp}} = \frac{\tau_{\bm{v}}C_kN_{k,\bm{v}_{k}}^{g}}{f_{k}} + \frac{\tau_{\bm{u}}C_kN_{k,\bm{u}_{k}}^{g}}{f_{k}},
\end{equation}
where $\tau_{\bm{u}}$ is the number of iterations for the global part updating, $N_{k,\bm{v}_k}^{g}$ is the size of the personalized part, $N_{k,\bm{u}_k}^{g}$ is the size of the global part, and $C_kN_{k,\bm{v}_{k}}^{g}$ and $C_kN_{k,\bm{u}_{k}}^{g}$ are the number of cycles to perform one local updating of the personalized and global parts, respectively.

\subsubsection{Uplink Transmission of Global Part}
After personalized and global parts updating in the $k$th edge device, the updated global part $\bm{u}_{k}^{g}$ is delivered to the edge server by uplink transmission. We consider an orthogonal frequency-division multiple access (OFDMA) protocol for uplink transmission \cite{9127160,9194337}. The transmission rate between the edge server and the $k$th device is expressed as
\begin{equation}
    R_{k,g}^{\text{up}} = {b}_{k}^{g}B\log_{2}\left(1 + \frac{h_{k}^{g}p_k}{\sigma^2}\right),
\end{equation}
where ${b}_{k}^{g}$ is the bandwidth fraction allocated to the $k$th mobile device, $B$ is the total bandwidth, $h_{k}^{g}$ is the channel gain, $p_k$ is the transmission power of the $k$th mobile device, and $\sigma^2$ is the noise power. Then, the uplink transmission latency is calculated as
\begin{equation}
    T_{k,g}^{\text{up}} = \frac{\hat{q}N_{k,\bm{u}_{k}}^{g}}{R_{k,g}^{\text{up}}},
\end{equation}
where $\hat{q}$ is the quantization bit.

\subsubsection{Global Part Aggregation}
When the edge server receives the global parts from edge devices, it averages them into a global part by
\begin{equation}
    \bm{u}^{g+1} = \frac{1}{|\mathcal{K}_{g}|}\sum_{k\in\mathcal{K}_{g}}\bm{u}_{k}^{g},
\end{equation}
where $\mathcal{K}_{g}$ is the set of edge devices participating in global part aggregation in the $g$th global communication round, and $|\mathcal{K}_{g}|$ is the number of edge devices.

Then, the edge server transmits the aggregated global part ${\bm{u}}^{g+1}$ to the edge devices for the next round of global and personalized parts updating. The edge server is usually equipped with a high computation unit, such as a graphics processing unit (GPU); thus, the latency of the global part aggregation is ignored.

\subsection{Local Computation and Uplink Latency}
Because of the high computation capability of the edge server and the sufficient bandwidth in downlink transmission, the computation latency of model aggregation in the edge server and downlink transmission latency of the global part are ignored. We mainly consider the latency of uplink transmission of the global part and local model updating, which is calculated as 
\begin{align}
    &T_{k}^{g} = T_{k,g}^{\text{cmp}} + T_{k,g}^{\text{up}}\\
    &= \frac{\tau_{\bm{v}}C_kN_{k,\bm{v}_{k}}^{g}}{f_{k}} + \frac{\tau_{\bm{u}}C_kN_{k,\bm{u}_{k}}^{g}}{f_{k}} +\frac{\hat{q}N_{k,\bm{u}_{k}}^{g}}{R_{k,g}^{\text{up}}}.
\end{align}
To be simplicity, synchronous uplink transmission of the global part and local model updating in edge devices are considered; thus, the computation and uplink transmission latency in each global communication is determined by the slowest one that finishes all global part transmission and local model updating, which is denoted as
\begin{equation}
    T_{g} = \max_{k\in\mathcal{K}}\{T_{k}^{g}\}.
\end{equation}

\subsection{Optimization Problem Formulation}
Given the previously mentioned system model, the objective of this work is to minimize the global loss function $F(\bm{u}, \bm{v}_{k})$ in (3) under the bandwidth and latency thresholds, and the optimization problem is formulated as
\begin{align}
    \min_{{b}_{k}^{g}}~~&F(\bm{u}, \bm{v}_{k}),\\
    s.t.~~&T_{g}\leq T_{\text{th}},\\
    ~~&\sum_{k=1}^{K}{b}_{k}^{g}\leq 1,\\
    ~~&{b}_{k}^{g}\in [0, 1].
\end{align}
In (16), $T_\text{th}$ represents the computation and communication latency constraint, and constraints in (17) and (18) represent the bandwidth fraction thresholds. To this end, we first consider partial model pruning to guarantee learning accuracy and decrease computation and communication latency. Then, we analyze the convergence bound of the proposed FL with partial model pruning and personalization algorithm and transform the optimization problem in (15) into optimizing the convergence bound.

\section{Convergence Analysis and Problem Transformation}
In this section, we start with model pruning of the global part. Then, we theoretically characterize the convergence bound of the proposed FL framework and transform the optimization problem into a joint pruning ratio and bandwidth allocation optimization.

\subsection{Model Pruning}
In wireless FL frameworks with partial model personalization, updating learning models on edge devices with limited computation capabilities and exchanging them between edge servers and devices under poor wireless channel conditions may lead to high computation and communication latency, even if only the global part $\bm{u}$ is shared among devices. We therefore consider pruning to reduce the size of the learning model.

Effectively pruning insignificant weights may reduce the model size without sacrificing learning performance. We consider an unstructured pruning method \cite{P_Molchanov}, whose the importance of weight is computed by a squared difference between prediction errors obtained with and without the corresponding model weight. For example, the importance of the $j$th model weight of the global part is calculated as
\begin{equation}
    I_{k,j} = \left(F_{k}(\bm{u}_k,\bm{v}_k) - F_{k}(\bm{u}_k,\bm{v}_k | u_{k,j} = 0)\right)^2.
\end{equation}
Weight importance increases as the corresponding error in (19) grows. Nevertheless, calculating $I_{k,j}$ for all model weights of the global part is time-consuming. To alleviate the computational burden, an alternative approach is to calculate the difference between the updated and non-updated $j$th model weight, denoted as
\begin{equation}
    \hat{I}_{k,j} = |\hat{u}_{k,j} - {u}_{k,j}|,
\end{equation}
where $\hat{u}_{k,j}$ and ${u}_{k,j}$ are the updated $j$th and non-updated $j$th model weight of the global part in the $k$th device, respectively. Calculating the importance using (20) is straightforward, as the updated global part weight $\hat{u}_{k,j}$ is already available through backpropagation.

\begin{figure}[!h]
    \centering
    \includegraphics[width=3.5 in]{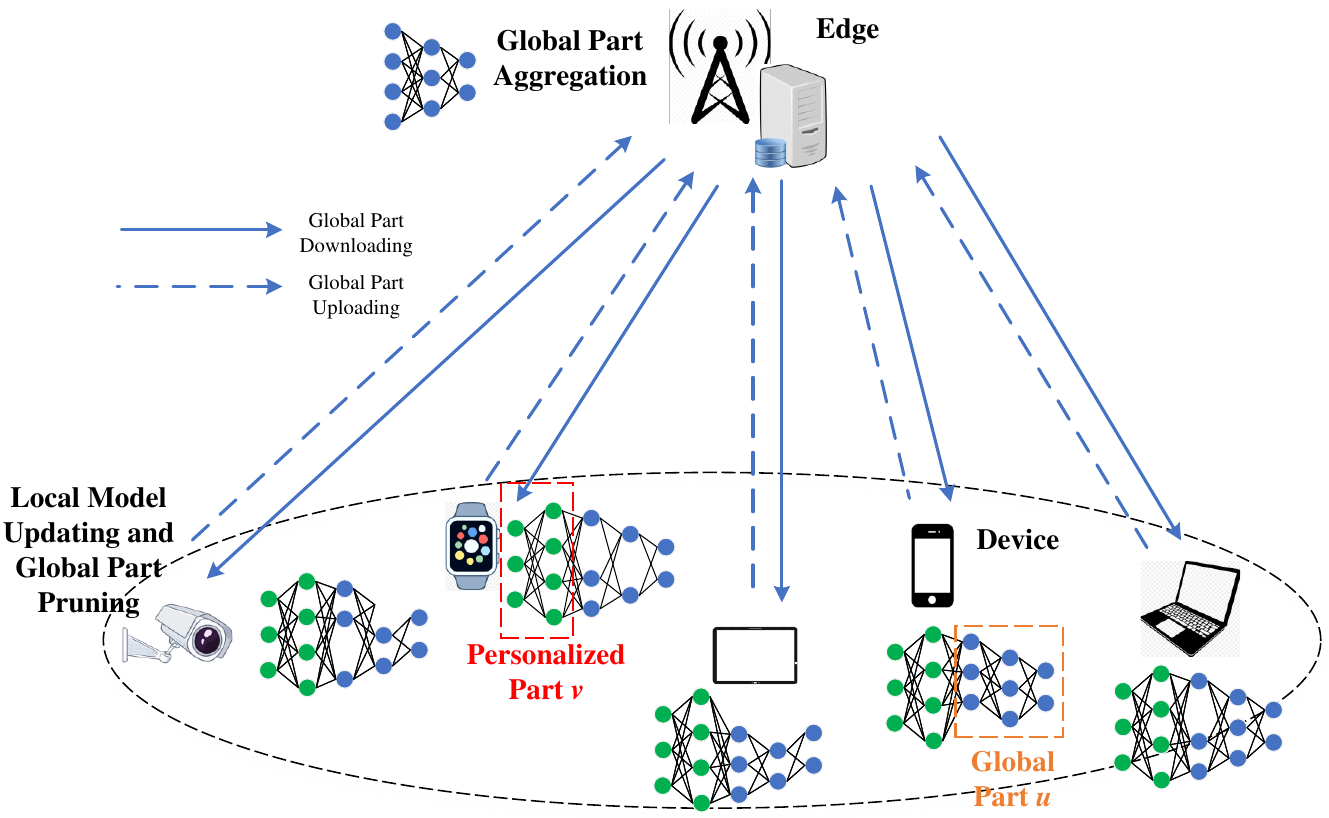}
    \caption{FL with partial model pruning and personalization framework.}
    \label{basic_modules}
\end{figure}

The proposed FL with partial model pruning and personalization is shown in Fig. 1, where the personalized parts
are only updated in local devices, and the global parts are
pruned and updated in local devices and aggregated in the
edge server. The learning process is updated as follows:

\subsubsection{Local Model Updating} Based on the local model updating in Section II.B, after $\tau_{\bm{v}}$ iterations of updating the personalized part $\bm{v}_{k}^{g,t}$, we begin updating the global part $\bm{u}_{k}^{g}$ of the $k$th device. The edge devices first update the global part $\bm{u}_{k}^{g}$ by $\bm{u}_{k}^{g} - \eta_{\bm{u}}\nabla F_{k}(\bm{u}_{k}^{g}, \bm{v}_{k}^{g,\tau_{\bm{v}}}, \xi_{k}^{g})$ for $\hat{\tau}_{\bm{u}}$ iterations, and $\hat{\tau}_{\bm{u}}$ can be a small value according to \cite{narang2017exploring,h.2018to}. Then, the importance of each weight in the global part is calculated by (20), and the weights of the global part are sorted in descending order. With a pruning ratio $\rho_{k}^{g}$, we deploy a pruning mask $\bm{m}_{k}^{g}$ to prune the global part $\bm{u}_{k}^{g}$, which is calculated as
\begin{equation}
    \bm{u}_{k}^{g,0} = \bm{u}^{g}\odot\bm{m}_{k}^{g}.
\end{equation}
In the pruning mask $\bm{m}_{k}^{g}$, if ${m}_{k}^{g,j} = 1$, $\bm{u}_{k}^{g,0}$ contains the $j$th model weight, otherwise, ${m}_{k}^{g,j} = 0$, and ${m}_{k}^{g,j}$ is determined by (20). In (21), the weights whose importance ranked last $\rho_{k}^{g}N_{k,\bm{u}_k}^{g}$ are pruned, and $N_{k,\bm{u}_k}^{g}$ is the size of the global part. Based on the updated personalized part $\bm{v}_{k}^{g,\tau_{\bm{v}}}$, the global part updating in the $t$th iteration is denoted as
\begin{equation}
    \bm{u}_{k}^{g,t+1} = \bm{u}_{k}^{g,t} - \eta_{\bm{u}}\nabla F_{k}(\bm{u}_{k}^{g,t}, \bm{v}_{k}^{g,\tau_{\bm{v}}}, \xi_{k}^{g,t})\odot\bm{m}_{k}^{g}.
\end{equation}
Furthermore, given the pruning ratio $\rho_{k}^{g}$, the number of local model weights after pruning is computed as
\begin{equation}
    N_{\rho_{k}^{g}} = N_{k,\bm{v}_k}^{g} + (1 - \rho_{k}^{g})N_{k,\bm{u}_k}^{g},
\end{equation}
where $N_{k,\bm{v}_k}^{g}$ and $N_{k,\bm{u}_k}^{g}$ are the size of the personalized and global parts, respectively.

Given the number of CPU cycles to update one model weight $C_k$ and the allocated CPU frequency of the $k$th device $f_{k}$, the computation latency of personalized and global parts updating $T_{k, g}^{\text{cmp}}$ in (8) is rewritten as 
\begin{equation}
    \hat{T}_{k, g}^{\text{cmp}} = \frac{\tau_{\bm{v}}C_kN_{k,\bm{v}_{k}}^{g}}{f_{k}} + \frac{\hat{\tau}_{\bm{u}}C_kN_{k,\bm{u}_{k}}^{g}}{f_{k}} + \frac{(1 - \rho_{k}^{g})\tau_{\bm{u}}C_kN_{k,\bm{u}_{k}}^{g}}{f_{k}},
\end{equation}
where $\tau_{\bm{u}}$ is the number of iterations for the global part updating, and $C_kN_{k,\bm{v}_{k}}^{g}$ and $C_kN_{k,\bm{u}_{k}}^{g}$ are the number of cycles to perform one local updating of the personalized and global parts, respectively. 

\subsubsection{Uplink Transmission of Global Part} According to the transmission rate calculated in (9), the uplink transmission latency of the global part in (10) is rewritten as
\begin{equation}
    \hat{T}_{k,g}^{\text{up}} = \frac{\hat{q}(1 - \rho_{k}^{g})N_{k,\bm{u}_{k}}^{g}}{R_{k,g}^{\text{up}}}.
\end{equation}

\subsubsection{Global Part Aggregation}
Given a pruning ratio $\rho_{k}^{g}$, some unimportant weights are pruned and not transmitted to the edge server.
We assume that the set of edge devices remaining the $j$th model weight of the global part is $\mathcal{K}_{g}^{j}$ and the number of edge devices containing the $j$th model weight is $|\mathcal{K}_{g}^{j}|$, thus, the aggregation of the $j$th model weight in the edge server is expressed
\begin{equation}
    \bm{u}^{g+1,j} = \frac{1}{|\mathcal{K}_{g}^{j}|}\sum_{k\in\mathcal{K}_{g}^{j}}\bm{u}_{k}^{g,j}.
\end{equation}

Furthermore, according to (12) and (13), the latency of uplink transmission of the global part and local model updating is rewritten as
\begin{align}
    &\hat{T}_{k}^{g} = \hat{T}_{k,g}^{\text{cmp}} + \hat{T}_{k,g}^{\text{up}}\\
    &= \frac{\tau_{\bm{v}}C_kN_{k,\bm{v}_{k}}^{g}}{f_{k}} + \frac{(1 - \rho_{k}^{g})\tau_{\bm{u}}C_kN_{k,\bm{u}_{k}}^{g}}{f_{k}} + \frac{\hat{\tau}_{\bm{u}}C_kN_{k,\bm{u}_{k}}^{g}}{f_{k}}\nonumber\\
    &+\frac{\hat{q}(1 - \rho_{k}^{g})N_{k,\bm{u}_{k}}^{g}}{R_{k,g}^{\text{up}}}.
\end{align}
Similarly, the synchronous uplink transmission of the global part and local model updating in edge devices is considered; thus, the computation and uplink transmission latency in each global communication in (14) is updated as
\begin{equation}
    \hat{T}_{g} = \max_{k\in\mathcal{K}}\{\hat{T}_{k}^{g}\}.
\end{equation}

\begin{figure}[!h]
    \centering
    \includegraphics[width=3.5 in]{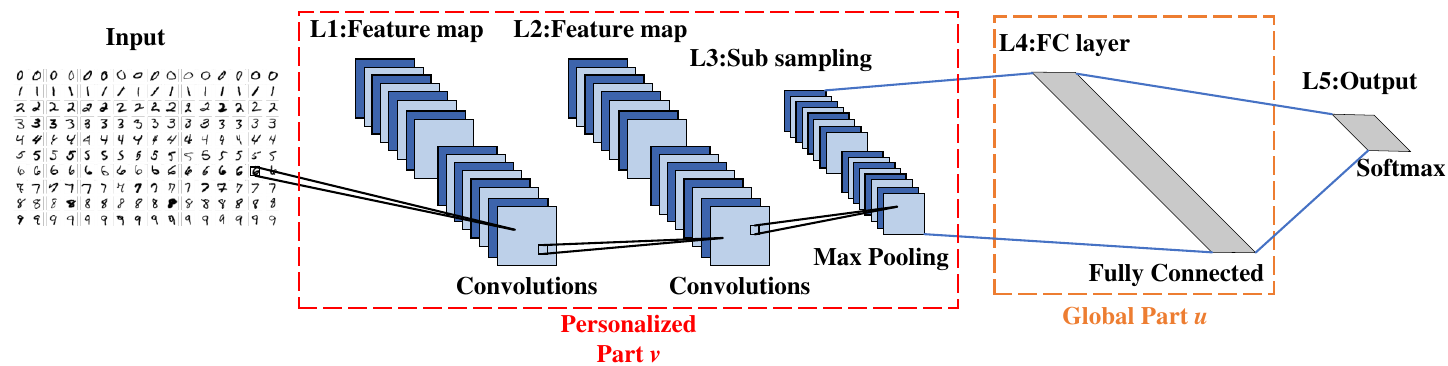}
    \caption{Selected personalized and global parts in CNN.}
    \label{basic_modules}
\end{figure}

To better illustrate the selected personalized and global parts in the proposed FL framework, we take a convolutional neural network (CNN) as an example, which is also used in experiment results. As shown in Fig. 2, CNN mainly includes an input layer, several convolution layers, several pooling layers, and a fully connected layer. In our work, the convolution layers and pooling layers are responsible for feature extraction, which belongs to the personalized parts. While the fully connected layers are used for pattern recognition, which belongs to the global parts. Model pruning in the convolution layers (personalized part) decreases the robust capability of CNN; thus, we mainly consider model pruning in the fully connected layers (global part).

\subsection{Proposed FL Convergence Analysis}
According to \cite{Ghadimi} and \cite{SShi}, the average $l_2$-norm of gradients is usually used to evaluate the convergence of the proposed FL framework. The following assumptions are used to facilitate convergence analysis of FL with partial model pruning and personalization.

$\textbf{Assumption 1.}$ All loss functions $F_{k}(\bm{u}, \bm{v}_k)$ are continuously differentiable with respect to $\bm{u}$ and $\bm{v}_k$, namely, $\nabla_{\bm{u}}F_{k}(\bm{u},\bm{v}_{k})$ is $L_{\bm{u}}$-Lipschitz continuous with $\bm{u}$, and $\nabla_{\bm{v}}F_{k}(\bm{u},\bm{v}_{k})$ is $L_{\bm{v}}$-Lipschitz continuous with $\bm{v}_{k}$, which are denoted as
\begin{equation}
    \|\nabla_{\bm{u}} F_{k}(\bm{u},\bm{v}_{k}) - \nabla_{\bm{u}} F_{k}(\hat{\bm{u}},\bm{v}_{k})\| \leq L_{\bm{u}}\|\bm{u} - \hat{\bm{u}}\|,
\end{equation}
and
\begin{equation}
    \|\nabla_{\bm{v}} F_{k}(\bm{u},\bm{v}_{k}) - \nabla_{\bm{v}} F_{k}(\bm{u},\hat{\bm{v}}_{k})\| \leq L_{\bm{v}}\|\bm{v}_{k} - \hat{\bm{v}}_{k}\|,
\end{equation}
where $L_{\bm{u}}$ and $L_{\bm{v}}$ are positive constants.

$\textbf{Assumption 2.}$ Global part pruning introduces noise and causes model error. Unlike the convergence analysis of FL with model personalization in \cite{pmlr-v162-pillutla22a} and \cite{konstantin}, based on \cite{xiaonan1} and \cite{Stich}, the upper bound of model error caused by $\rho_{k}^{g}$ is expressed as 
\begin{equation}
    \mathbb{E}\|\bm{u}_{k}^{g} - \bm{u}_{k}^{g}\odot\bm{m}_{k}^{g}\|^2\leq\rho_{k}^{g}D^2,
\end{equation}
where $D$ is a positive constant.

$\textbf{Assumption 3.}$ According to \cite{Bartlett} and \cite{Salimans}, the upper bound of the second moments of stochastic gradients of global and personalized parts are denoted as
\begin{equation}
    \mathbb{E}\|\nabla_{\bm{u}} F_{k}(\bm{u}_{k}^{g,t}, \bm{v}_{k}^{g,t}, \xi_{k}^{g,t})\|^2\leq \phi_{\bm{u}}^2,
\end{equation}
and
\begin{equation}
    \mathbb{E}\|\nabla_{\bm{v}} F_{k}(\bm{u}_{k}^{g,t}, \bm{v}_{k}^{g,t}, \xi_{k}^{g,t})\|^2\leq \phi_{\bm{v}}^2,
\end{equation}
respectively. In (33) and (34), $\phi_{\bm{u}}$ and $\phi_{\bm{v}}$ are positive constants, and $\xi_{k}^{g,t}$ is the set consisting of mini-batch data samples for any $k,g,t$.

$\textbf{Assumption 4.}$ The stochastic gradients of global and personalized parts are unbiased and have bounded variance, which is denoted as
\begin{equation}
    \mathbb{E}[\nabla_{\bm{u}} F_{k}(\bm{u}_{k}^{g,t}, \xi_{k}^{g,t})] = \nabla_{\bm{u}} F_{k}(\bm{u}_{k}^{g,t}),
\end{equation}
and
\begin{equation}
    \mathbb{E}[\nabla_{\bm{v}} F_{k}(\bm{v}_{k}^{g,t}, \xi_{k}^{g,t})] = \nabla_{\bm{v}} F_{k}(\bm{v}_{k}^{g,t}),
\end{equation}
respectively. Furthermore, there exist constants $\hat{\sigma}_{\bm{u}}$ and $\hat{\sigma}_{\bm{v}}$ satisfying
\begin{equation}
    \mathbb{E}\|\nabla_{\bm{u}} F_{k}(\bm{u}_{k}^{g,t}, \xi_{k}^{g,t}) - \nabla_{\bm{u}} F_{k}(\bm{u}_{k}^{g,t})\|^2\leq \hat{\sigma}_{\bm{u}}^2,
\end{equation}
and
\begin{equation}
    \mathbb{E}\|\nabla_{\bm{v}} F_{k}(\bm{v}_{k}^{g,t}, \xi_{k}^{g,t}) - \nabla_{\bm{v}} F_{k}(\bm{v}_{k}^{g,t})\|^2\leq \hat{\sigma}_{\bm{v}}^2,
\end{equation}
respectively.

$\textbf{Assumption 5.}$ There exist $\delta\geq 0$ and $\varphi\geq 0$ for all $\bm{u}$ and $\bm{V} = \sum_{k=1}^{K}\bm{v}_{k}$ satisfying
\begin{align}
    \frac{1}{K}\sum_{k=1}^{K}\|\nabla_{\bm{u}}F_{k}(\bm{u},\bm{v}_{k})-\nabla_{\bm{u}}F(\bm{u},\bm{V})\|^2\nonumber\\
    \leq \delta^2 + \varphi^{2}\|\nabla_{\bm{u}}F(\bm{u},\bm{V})\|^2.
\end{align}
Partial gradient diversity characterizes how local steps on one device affect convergence globally.

\textbf{Theorem 1:} Based on the aforementioned assumptions, the upper bound of the convergence rate of the proposed FL is derived as:
\begin{align}
    &\frac{1}{G}\sum_{g=1}^{G}\left[\frac{\eta_{\bm{v}}\tau_{\bm{v}}}{8}\sum\limits_{k=1}^{K}\left\|\nabla_{\bm{v}}F_{k}(\bm{u}^{g},\bm{v}_{k}^{g})\right\|^2 \right.\nonumber\\
    &+ \frac{\eta_{\bm{u}}\tau_{\bm{u}}}{2}\sum_{j=1}^{N}\left\|\nabla_{\bm{u}} F^{j}(\bm{u}^{g},\bm{V}^{g+1})\right\|^2\Bigg]  \nonumber\\
    &\leq
     \frac{\mathbb{E}[F(\bm{u}^{0}, \bm{V}^{0}) - F(\bm{u}^{*},\bm{V}^{*})]}{G}+ A_1  + A_2\sum_{g=1}^{G}\sum_{k=1}^{K} \rho_{k}^{g},
\end{align}
where $A_1$ and $A_2$ are expressed as
\begin{align}
    &A_1 = \frac{\eta_{\bm{v}}^{2}\tau_{\bm{v}}^{2}\hat{\sigma}_{\bm{v}}^{2}L_{{\bm{v}}}}{2} + 4\eta_{\bm{v}}^{3}L_{{\bm{v}}}^2\hat{\sigma}_{\bm{v}}^{2}\tau_{\bm{v}}^{2}(\tau_{\bm{v}}-1)+\frac{3\eta_{\bm{u}}^2N^2\tau_{\bm{u}}^2\phi_{\bm{u}}^2L_{\bm{u}}}{2} \nonumber\\
    &\!+\!\frac{N\phi_{\bm{u}}^2K\eta_{\bm{u}}^3L_{\bm{u}}^2\tau_{\bm{u}}^3\!+\!3N^2\eta_{\bm{u}}^2\tau_{\bm{u}}^2K\hat{\sigma}_{\bm{u}}^2L_{\bm{u}} \!+ \!3N^2L_{\bm{u}}^3\tau_{\bm{u}}^4\eta_{\bm{u}}^4K\phi_{\bm{u}}^2}{2\kappa^{*}},
\end{align}
and
\begin{equation}
    A_2 = \frac{N\eta_{\bm{u}} \tau_{\bm{u}}L_{\bm{u}}^2D^2+3N^2\eta_{\bm{u}}^2 L_{\bm{u}}^3D^2\tau_{\bm{u}}^2} {G\kappa^{*}}.
\end{equation}
In (40), (41), and (42), $N$ is the total number of model weights of the global part, and $\kappa^{*}$ is the minimum occurrence of the model weights in the global part in all communication rounds. Intuitively, if a model weight
is pruned and never included in any local models, it is impossible for it to be updated. 

$\mathit{Proof:}$ Please refer to Appendix.

$\mathit{Remark~1:}$ Based on Theorem 1, we can obtain that given the number of iterations of both the global and personalized parts, the number of global communication rounds, and the learning rate, the upper bound of the convergence rate of the proposed FL is mainly determined by the pruning ratio. Therefore, we mainly consider optimizing the pruning ratio under the latency and bandwidth constraints.

\subsection{Optimization Problem Transformation}
Motivated by Theorem 1, we need to maximize the convergence rate, and the optimization problem is reformulated as
\begin{align}
    \min_{{b}_{k}^{g}, \rho_{k}^{g}}~~&A_2\sum_{g=1}^{G}\sum_{k=1}^{K} \rho_{k}^{g},\\
    s.t.~~&\hat{T}_{g}\leq T_{\text{th}},\\
    ~~&(17), (18) \nonumber\\
    ~~&\rho_{k}^{g}\in [0, 1].
\end{align}
The constraint in (45) represents the threshold of the pruning ratio. To minimize the global loss function, a proper pruning ratio of the global part and a proper bandwidth fraction of each edge device should be carefully selected based on the constraints in (17), (18), (44) and (45).

\section{Pruning Ratio and Bandwidth Fraction Optimization}
To optimize the pruning ratio and bandwidth fraction, the optimization problem in (43) is divided into two sub-problems.

\subsection{Pruning Ratio Optimization}
According to the constraint in (44), the computation and uplink transmission latency of the $k$th edge device should 
not larger than the latency constraint, which is written as
\begin{equation}
    \frac{\tau_{\bm{v}}C_kN_{k,\bm{v}_{k}}^{g}}{f_{k}} + (1 - \rho_{k}^{g})\left(\frac{\tau_{\bm{u}}C_kN_{k,\bm{u}_{k}}^{g}}{f_{k}} + \frac{\hat{q}N_{k,\bm{u}_{k}}^{e}}{R_{k,g}^{\text{up}}}\right)\leq T_{\text{th}}.
\end{equation}

\textbf{Theorem 2:}
Based on (46), the lower bound of the pruning ratio in the $g$th global communication round should satisfy
\begin{equation}
    \rho_{k}^{g*} \geq\left(1 - \frac{T_{\text{th}} - T_{k,g}^{\text{cmp-Per}}}{T_{k,g}^{\text{cmp-G}} + T_{k,g}^{\text{com-G}}}\right)^{+},
\end{equation}
where $T_{k,g}^{\text{cmp-Per}}$ is the computation latency of the personalized part, and $T_{k,g}^{\text{com-G}}$ and $T_{k,g}^{\text{cmp-G}}$ are transmission and computation latency of the global part, respectively, where $(s)^{+} = \max(s,0)^{+}$.

\begin{proof}
The latency constraint in (46) is rewritten as
\begin{equation}
    T_{k,g}^{\text{cmp-Per}} + (1-\rho_{k}^{g})(T_{k,g}^{\text{com-G}} + T_{k,g}^{\text{com-G}})\leq T_{\text{th}},
\end{equation}
where $T_{k,g}^{\text{cmp-Per}} = \frac{\tau_{\bm{v}}C_kN_{k,\bm{v}_{k}}^{g}}{f_{k}}$, $T_{k,g}^{\text{com-G}}=\frac{\tau_{\bm{u}}C_kN_{k,\bm{u}_{k}}^{g}}{f_{k}}$, and $T_{k,g}^{\text{com-G}} = \frac{\hat{q}N_{k,\bm{u}_{k}}^{g}}{R_{k,g}^{\text{up}}}$. Then, $\rho_{k}^{g}$ is derived as (47).
\end{proof}

$\mathit{Remark~2:}$ Based on Theorem 2 and (48), both the computation capability and uplink transmission rate are able to determine the pruning ratio of each device and a small pruning ratio is adopted for the device with a high uplink transmission rate and computation capability.

\subsection{Bandwidth Fraction Optimization}
Based on the lower bound of the pruning ratio in (47), (43) is rewritten as
\begin{equation}
    \min_{{b}_{k}^{g}}~~A_2\sum_{g=1}^{G}\sum_{k=1}^{K} \left(1 - \frac{T_{\text{th}} - T_{k,g}^{\text{cmp-Per}}}{T_{k,g}^{\text{cmp-G}} + T_{k,g}^{\text{com-G}}}\right),
\end{equation}
with the constraints (17) and (18). Given $T_{k,g}^{\text{com-G}} = \frac{\hat{q}N_{k,\bm{u}_{k}}^{g}}{R_{k,g}^{\text{up}}}$, (49) is further rewritten as 
\begin{equation}
    \min_{{b}_{k}^{g}}~~A_2\sum_{g=1}^{G}\sum_{k=1}^{K} \left(1 - \frac{R_{k,g}^{\text{up}}(T_{\text{th}} - T_{k,g}^{\text{cmp-Per}})}{R_{k,g}^{\text{up}}T_{k,g}^{\text{cmp-G}} + \hat{q}N_{k,\bm{u}_{k}}^{g}}\right).
\end{equation}
To achieve the closed-form solution of bandwidth fraction $b_{k}^{g}$, we first prove that the optimization problem in (50) is convex concerning the bandwidth fraction $b_{k}^{g}$ in Lemma 1.

$\textbf{Lemma~1:}$ The optimization problem in (50) is convex concerning the bandwidth fraction.

\begin{proof}
The optimization problem in (50) is equal to 
\begin{equation}
    F(b) = \sum_{g=1}^{G}\sum_{k=1}^{K}f(b_{k}^{g}) = \sum_{g=1}^{G}\sum_{k=1}^{K}\left(1 - \frac{b_{k}^{g}V_1}{b_{k}^{g}V_2 + V_3}\right),
\end{equation}
where $V_1, V_2, V_3 > 0$, and $0\leq b_{k}^{g}\leq 1$. To prove Lemma 1, we need to analyze the convexity of the function $f(b_{k}^{g})$. The first derivative of $f(b_{k}^{g})$ is computed as
\begin{equation}
    f^{'}(b_{k}^{g}) = -\frac{V_1V_3}{(b_{k}^{g}V_2 + V_3)^2}.
\end{equation}
Then, the second derivative of $f(b_{k}^{g})$ is derived as
\begin{equation}
    f^{''}(b_{k}^{g}) = \frac{2V_1V_2V_3}{(b_{n}^{e}V_2 + V_3)^3} > 0.
\end{equation}
As a result, the optimization problem in (50) is convex with respect to the bandwidth fraction. 
\end{proof}

Then, the optimal bandwidth fraction of each device is calculated by the Lagrange multiplier, as shown in the following theorem.

\textbf{Theorem 3:} The closed-form solution of bandwidth fraction of the $k$th device is derived as
\begin{equation}
    b_{k}^{g*} = \frac{\sqrt{\frac{(T_{\text{th}} - T_{k,g}^{\text{cmp-Per}})\hat{q}N_{k,\bm{u}_{k}}^{g}B\log_{2}\left(1 + \frac{h_{k}^{g}p_k}{\sigma^2}\right)}{\lambda^{*}}}-\hat{q}N_{k,\bm{u}_{k}}^{g}}{B\log_{2}\left(1 + \frac{h_{k}^{g}p_k}{\sigma^2}\right)T_{k,g}^{\text{cmp-G}}},
\end{equation}
where $\lambda^{*}$ is the optimal Lagrange multiplier.

\begin{figure*}[ht]
	\centering
	\subfloat[]{\includegraphics[width=2.1in]{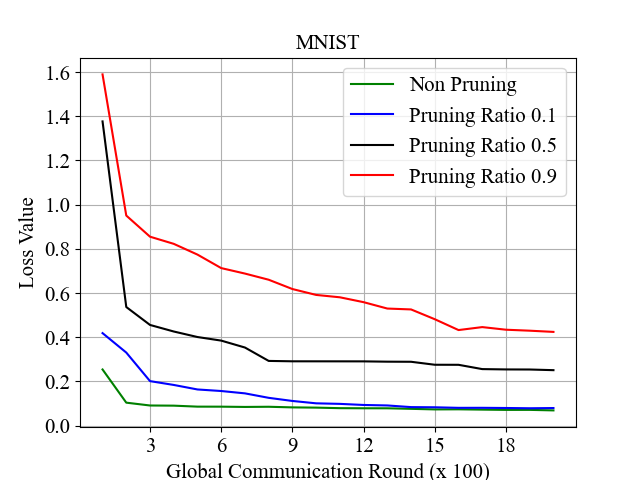}}\label{fig_first_case}
	\hfil
	\subfloat[]{\includegraphics[width=2.1in]{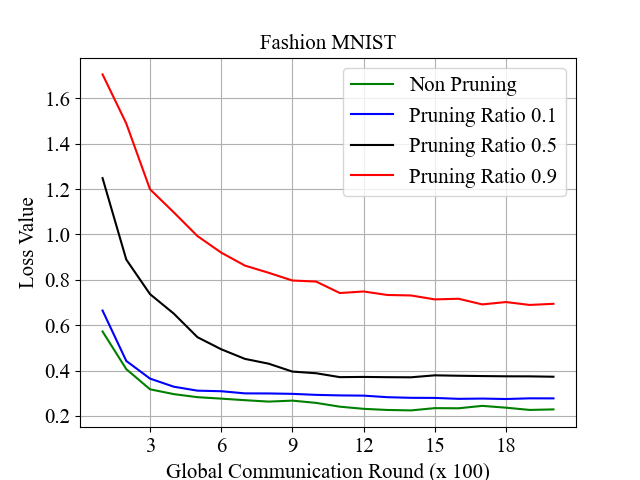}}\label{fig_second_case}
    \hfil
	\subfloat[]{\includegraphics[width=2.0in]{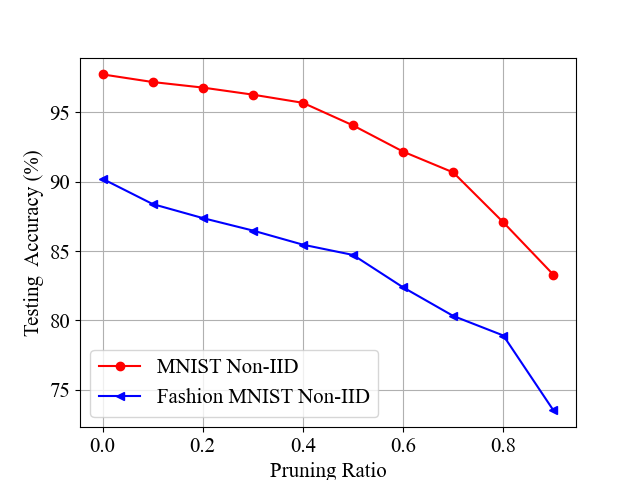}}\label{fig_third_case}
	\caption{(a) The relationship between the loss value of the proposed FL and pruning ratios on the non-IID MNIST dataset. (b) The relationship between the loss value of the proposed FL and pruning ratios on the non-IID Fashion MNIST dataset. (c) The performance of the testing accuracy with increasing pruning ratio of the proposed FL.}
	\label{basic_modules}
\end{figure*}

\begin{proof}
According to the objective function in (50) and the threshold in (17), the Lagrange function is written as
\begin{align}
    \mathcal{L}(b_{k}^{g},\lambda) = &\sum_{k=1}^{K}\left(1 - \frac{{b}_{k}^{g}B\log_{2}\left(1 + \frac{h_{k}^{g}p_k}{\sigma^2}\right)(T_{\text{th}} - T_{k,g}^{\text{cmp-Per}})}{{b}_{k}^{g}B\log_{2}\left(1 + \frac{h_{k}^{g}p_k}{\sigma^2}\right)T_{k,g}^{\text{cmp-G}} + \hat{q}N_{k,\bm{u}_{k}}^{g}}\right)\nonumber\\
    &+\lambda\left(\sum_{k=1}^{K}b_{k}^{g}-1\right).
\end{align}
In (55), $\lambda$ is a Lagrange multiplier. Then, we consider the Karush-Kuhn-Tucker (KKT) conditions to solve the problem, which is written as
\begin{equation}
    \frac{\partial\mathcal{L}}{\partial b_{k}^{g}} = \lambda - \frac{(T_{\text{th}} - T_{k,g}^{\text{cmp-Per}})\hat{q}N_{k,\bm{u}_{k}}^{g}B\log_{2}\left(1 + \frac{h_{k}^{g}p_k}{\sigma^2}\right)}{\left[{b}_{k}^{g}B\log_{2}\left(1 + \frac{h_{k}^{g}p_k}{\sigma^2}\right)T_{k,g}^{\text{cmp-G}} + \hat{q}N_{k,\bm{u}_{k}}^{g}\right]^2}=0,
\end{equation}
\begin{equation}
    \lambda\left(\sum_{k=1}^{K}b_{k}^{g} - 1\right) = 0,~\lambda\geq 0.
\end{equation}
According to KKT conditions, the optimal bandwidth allocation for each device is obtained as Theorem 3.
\end{proof}

By plugging (54) into (47), the lower bound of the pruning ratio of the $k$th device is calculated as
\begin{equation}
    \rho_{k}^{g*} = 1 - \frac{b_{k}^{g*}(T_{\text{th}} - T_{k,g}^{\text{cmp-Per}})B\log_{2}\left(1 + \frac{h_{k}^{g}p_k}{\sigma^2}\right)}{b_{k}^{g*}T_{k,g}^{\text{cmp-G}}B\log_{2}\left(1 + \frac{h_{k}^{g}p_k}{\sigma^2}\right) + \hat{q}N_{k,\bm{u}_{k}}^{g}}.
\end{equation}

$\mathit{Remark~3:}$ According to Theorem 3, we can observe that the devices with bad channel conditions are allocated with more bandwidth to satisfy transmission latency. In addition, the device with high computation capacity is allocated with more bandwidth, improving the convergence rate and decreasing computation latency. Detailed procedures of the proposed FL framework are shown in \textbf{Algorithm 1}.

\begin{algorithm}[t]
\begin{algorithmic}[1]
\caption{FL with partial model pruning and personalization}
\STATE Local dataset $\mathcal{D}_k$ of the $k$th edge device, number of global communication rounds $G$, number of local iterations of the global and personalized parts $\tau_{\bm{u}}$, $\hat{\tau}_{\bm{u}}$ and $\tau_{\bm{v}}$, learning rate $\eta_{\bm{u}}$ and $\eta_{\bm{v}}$ of the global and personalized parts, global model parameterized by $\bm{u}^{g}$, local model parameterized by $\bm{u}_{k}^{g,t}$ and $\bm{v}_{k}^{g,t}$.
\FOR{global communication round $g$ = 1,...,$G$}
    \FOR{edge device $k$ = 1,...,$K$}
        \FOR{iteration for updating personalized part $t$ = 1,2,...,$\tau_{\bm{v}}$}
            \STATE Update personalized part $\bm{v}_{k}^{g,t}$ as (6).
        \ENDFOR
        \STATE Update the global part $\bm{u}_{k}^{g}$ by $\bm{u}_{k}^{g} - \eta_{\bm{u}}\nabla F_{k}(\bm{u}_{k}^{g}, \bm{v}_{k}^{g,\tau_{\bm{v}}}, \xi_{k}^{g})$ for $\hat{\tau}_{\bm{u}}$ iterations, calculate the optimal bandwidth fraction $b_{k}^{g*}$ by (54), compute the pruning ratio $\rho_{k}^{g*}$ by (58), and generate pruning mask $\bm{m}_{k}^{g}$ by (20).
        \STATE Initialize $\bm{u}_{k}^{g,0} = \bm{u}^{e}\odot\bm{m}_{k}^{e}$.
        \FOR{iteration for updating global part $t$ = 1,2,...,$\tau_{\bm{u}}$}
            \STATE Update the global part $\bm{u}_{k}^{g,t}$ as (22).
        \ENDFOR
    \ENDFOR
    \STATE Transmit the pruned global part to the edge server based on the bandwidth fraction $b_{k}^{g*}$ calculated by (54).
    \FOR {model weight $j$ in global part $\bm{u}_{k}^{g,\tau_{\bm{u}}}$}
        \STATE Find $\mathcal{K}_{g}^{j} = \{k: m_{k}^{g,j} = 1\}$.
        \STATE Update $\bm{u}^{g,j}$ as (26).
    \ENDFOR
\ENDFOR
\end{algorithmic}
\end{algorithm}

\section{Experiment Results}
This section examines our proposed FL framework. The experiment considers one edge server providing ten devices with wireless and global part aggregation services. The bandwidth allocated to the edge server is $20~\text{MHz}$, the transmission power of the device is $28~\text{dBm}$, the CPU frequency of the device is $3~\text{GHz}$, the quantization bit is $32$, the latency threshold is $25~\text{ms}$, and the AWGN noise power is $-110~\text{dBm}$. For the learning model, we consider CNN for image classification over MNIST and Fashion MNIST datasets. We set the convolutional layers and fully connected layers to be personalized and global parts, respectively. Both the learning rate of the personalized and global parts is $0.001$. The mini-batch size is $128$. Wireless channels transmit the global parts between the edge server and devices. The input size of CNN is 1 $\times$ 28 $\times$ 28, and the sizes of the first and second
convolutional layers are 32 $\times$ 28 $\times$ 28 and 64 $\times$ 14 $\times$ 14, respectively. The sizes of the first and second max-pooling layers are 32×14×14 and 64 $\times$ 7 $\times$ 7, respectively. The sizes of the first and
second fully connected layers are 3136 and 128, respectively. The size of the output layer is 10. Each device exclusively has data from several labels.

\begin{figure}[!h]
    \centering
    \includegraphics[width=3.5 in]{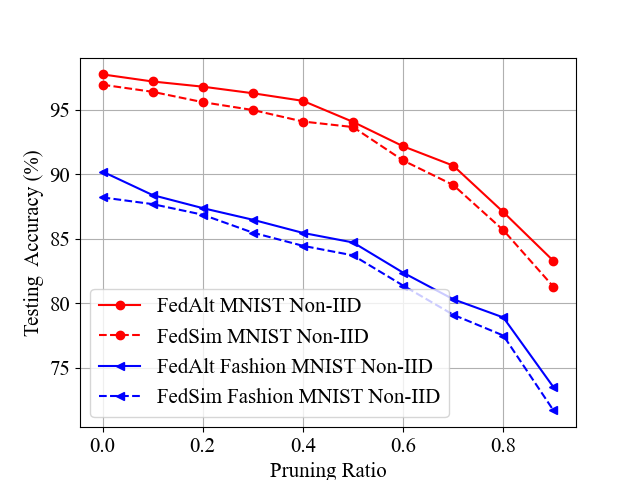}
    \caption{Comparison of the testing accuracy of the proposed FL by alternatively and simultaneously local updating.}
    \label{basic_modules}
\end{figure}

\begin{figure*}[ht]
	\centering
	\subfloat[]{\includegraphics[width=2.1in]{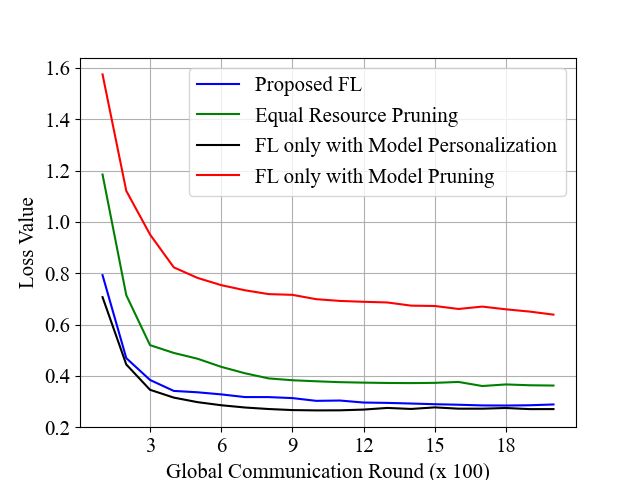}}\label{fig_first_case}
	\hfil
	\subfloat[]{\includegraphics[width=2.1in]{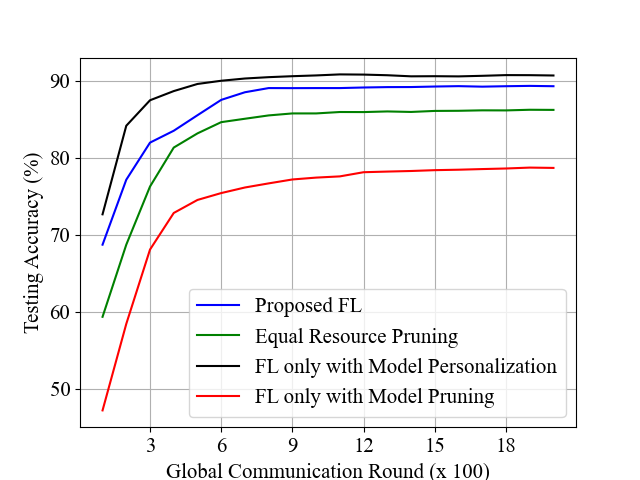}}\label{fig_second_case}
    \hfil
	\subfloat[]{\includegraphics[width=2.1in]{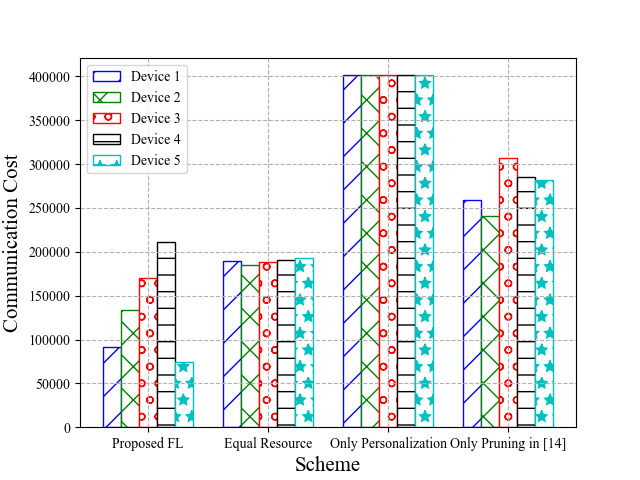}}\label{fig_third_case}
	\caption{(a) Loss value comparison of joint optimization of pruning ratio and bandwidth fraction of the proposed FL framework with other three FL schemes. (b) Testing accuracy comparison of joint optimization of pruning ratio and bandwidth fraction of the proposed FL framework with other three FL schemes. (c) Comparison of communication costs on different FL schemes.}
	\label{basic_modules}
\end{figure*}

\begin{figure*}[ht]
	\centering
	\subfloat[]{\includegraphics[width=2.1in]{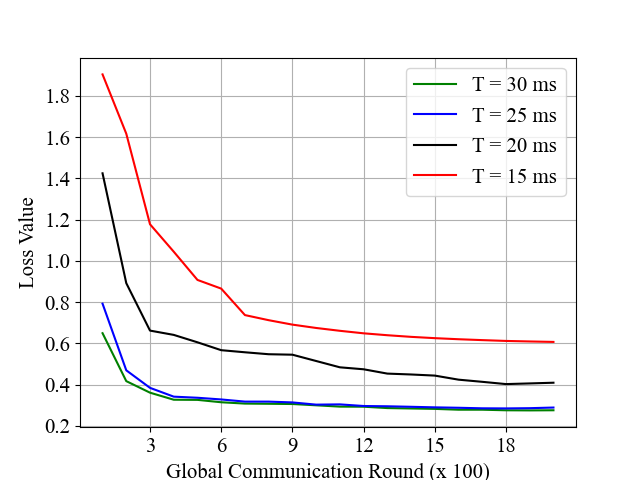}}\label{fig_first_case}
	\hfil
	\subfloat[]{\includegraphics[width=2.1in]{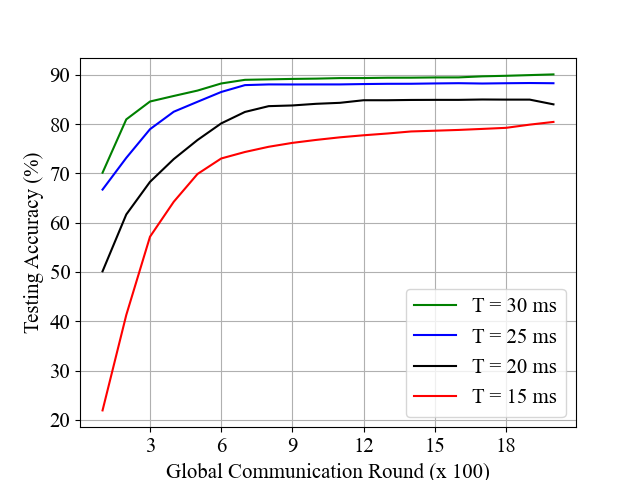}}\label{fig_second_case}
    \hfil
	\subfloat[]{\includegraphics[width=2.1in]{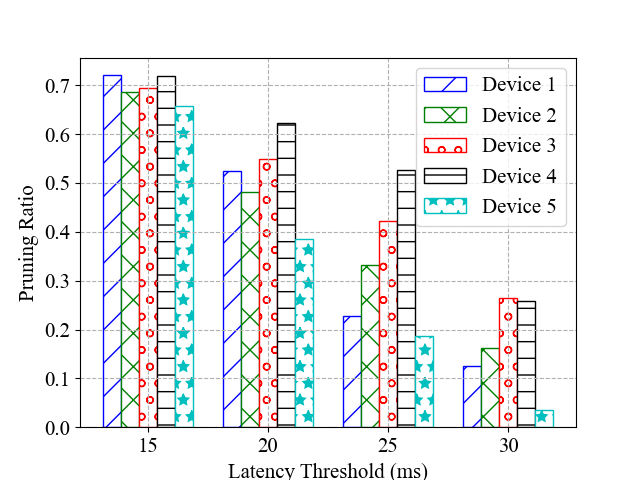}}\label{fig_third_case}
	\caption{(a) Loss value of the proposed FL framework with different latency thresholds. (b) Testing accuracy of the proposed FL framework with different latency thresholds. (c) The selected pruning ratio to achieve a given latency constraint.}
	\label{basic_modules}
\end{figure*}

\subsection{FL with Partial Model Pruning and Personalization}
The relationship between the loss value of the proposed FL and pruning ratios on non-IID MNIST and Fashion MNIST datasets is shown in Fig. 3 (a) and Fig. 3 (b), respectively. The performance of the testing accuracy with increasing pruning ratio of the proposed FL is shown in Fig. 3 (c). We find that the convergence rate decreases, and the value loss increases with the increasing pruning ratio. This is because more model weights are pruned with a higher pruning ratio, which leads to a lower learning ability. The comparison of the testing accuracy of the proposed FL by alternatively (FedAlt) and simultaneously (FedSim) local updating is plotted in Fig. 4. We observe that the testing accuracy of FedAlt is a bit higher than that of FedSim.

\begin{table*}
\centering
\caption{Computation and communication latency of each global communication round (ms)}
\begin{tabular}[c]{c|c|c|c|c}
\hline
\hline Scheme & Proposed FL & Equal Resource Pruning & Partial Personalization in \cite{pmlr-v162-pillutla22a} & Partial Pruning in \cite{xiaonan1}\\
\hline FL with Model Pruning and Personalization & 25 & 38 $\pm$ 0.45 & 55 $\pm$ 0.55  & 45\\

\hline
\hline
\end{tabular}
\end{table*}

\subsection{Proposed FL over Wireless Networks}
This section presents the performance of the joint design of pruning ratio and bandwidth fraction allocation for the proposed FL and how the latency threshold affects the proposed FL.

\subsubsection{Pruning Ratio and Bandwidth Fraction Optimization} Three baselines are considered to compare with the proposed FL. These three baselines are presented as follows.
\begin{itemize}
    \item $\textbf{Equal Resource Pruning: }$ The pruning ratio is optimized based on the equally allocated bandwidth. 
    \item $\textbf{FL only with Model Personalization in \cite{pmlr-v162-pillutla22a}: }$ In \cite{pmlr-v162-pillutla22a}, only FL with model personalization is considered. 
    \item $\textbf{FL only with Model Pruning in \cite{xiaonan1}: }$ In \cite{xiaonan1}, only FL with optimal model pruning was considered.
\end{itemize}

Table I shows the model updating and transmission latency of each global communication round of these four schemes mentioned above. It is shown that the computation and communication latency of the proposed FL is much smaller than that of the FL only with model pruning in \cite{xiaonan1} or the FL only with model personalization in \cite{pmlr-v162-pillutla22a}. This is because only the pruned global part needs to be transmitted between the edge server and devices. Also, it is observed that the latency of the proposed FL is smaller than that of the equal resource pruning scheme. This is because the proposed FL is able to select the optimal pruning ratio of the global part based on the allocated bandwidth, which further reduces the latency. 

The performance comparison of joint optimization of pruning and bandwidth fraction optimization of the proposed FL with the other three baselines on the non-IID Fashion MNIST dataset is presented in Fig. 5 (a) and Fig. 5 (b), respectively. It is observed that the loss value and testing accuracy of the proposed FL framework are better than the equal resource pruning scheme and FL only with model pruning in \cite{xiaonan1}. This is because the joint design of pruning ratio and bandwidth allocation by KKT conditions is able to prune the unimportant weights based on dynamic wireless environments, and partial model personalization enables learning the data heterogeneity of different devices. Also, from Fig. 5 (a) and Fig. 5 (b), the learning performance of the proposed FL 
is close to FL only with partial model personalization. However, as shown in Fig. 5 (c), the communication cost is about $50\%$ less than FL only with model personalization. The communication cost quantifies the number of model weights of the global part required to be delivered for model aggregation. This is because only the pruned global part needs to be transmitted to the edge server for model aggregation without sacrificing the learning performance.  

\subsubsection{Effect of Latency Threshold} Four latency thresholds ($15\text{ms}$, $20\text{ms}$, $25\text{ms}$, and $30\text{ms}$) are considered to show the effects on the proposed FL on the non-IID Fashion MNIST dataset in Fig. 6 (a) and Fig. 6 (b), respectively. Fig. 6 (c) shows the optimized pruning ratio calculated by KKT conditions satisfying the given latency constraints. It is obtained that with increasing latency thresholds, the loss value decreases, and the testing accuracy increases. Also, the number of global communication rounds achieving convergence becomes smaller with increasing latency thresholds. This is because a small pruning ratio is selected with a large latency threshold, and more important model weights of the global part are kept to improve the learning ability.

\section{Conclusions}
In this paper, a computation and communication-efficient FL framework with partial model pruning and personalization over wireless networks was proposed to adapt to data heterogeneity and dynamical wireless environments. Particularly, an upper bound on the $l_2$-norm of gradients for the proposed FL frameworks was derived. Subsequently, based on the convergence analysis of the proposed FL, the closed-form solutions of pruning ratio and bandwidth allocation were derived to maximize the convergence rate under latency and bandwidth thresholds by KKT conditions. Simulation results have demonstrated that our proposed FL framework achieved similar learning accuracy compared to FL only with partial model personalization and reduced about $50\%$ computation and communication overheads.

\appendix
\section{Appendix}
This section presents the detailed convergence of the proposed FL, and we use the following inequalities throughout the proof.

According to Jensen's inequality, for any $\bm{s}_{e}\in\mathbb{R}^{d}, e\in\{1, 2, ..., E\}$, we have
\begin{equation}
    \left\|\frac{1}{E}\sum_{e=1}^{E}\bm{s}_{e}\right\|^2\leq \frac{1}{E}\sum_{e=1}^{E}\|\bm{s}_{e}\|^2,
\end{equation}
and directly achieves
\begin{equation}
    \left\|\sum_{e=1}^{E}\bm{s}_{e}\right\|^2\leq E\sum_{e=1}^{E}\|\bm{s}_{e}\|^2.
\end{equation}
Peter-Paul inequality is shown as
\begin{equation}
\langle\bm{s}_1,\bm{s}_2\rangle \leq \frac{1}{2}\|\bm{s}_1\|^2 + \frac{1}{2}\|\bm{s}_2\|^2,
\end{equation}
and for any constant $m>0$ and $\bm{s}_{1}, \bm{s}_{2}\in\mathbb{R}^{d}$, we have
\begin{equation}
    \|\bm{s}_{1} + \bm{s}_{2}\|^2\leq (1+m)\|\bm{s}_{1}\|^2 + \left(1 + \frac{1}{m}\right)\|\bm{s}_{2}\|^2.
\end{equation}

\textbf{Proof of the Convergence: } In the FL with partial model pruning and personalization, the objective is to minimize the function
\begin{equation}
    F(\bm{u}, \bm{V}) = \frac{1}{K}\sum_{k=1}^{K}F_{k}(\bm{u}, \bm{v}_{k}),
\end{equation}
where $\bm{V} = (\bm{v}_{1},...,\bm{v}_{K})$ is a concatenation of all personalized parts. We use $L$-Lipschitz in Assumption 1 to give convergence analysis. We begin with
\begin{align}
    F(\bm{u}^{g+1}, \bm{V}^{g+1}) &- F(\bm{u}^{g},\bm{V}^{g}) = F(\bm{u}^{g}, \bm{V}^{g+1}) - F(\bm{u}^{g},\bm{V}^{g}) \nonumber\\
    &+ F(\bm{u}^{g+1}, \bm{V}^{g+1}) - F(\bm{u}^{g},\bm{V}^{g+1}).
\end{align}
In (64), the first line corresponds to the effect of the $\bm{v}$-step and is easy to obtain the upper bound with standard techniques. According to \cite{pmlr-v162-pillutla22a}, we have
\begin{align}
    &\mathbb{E}\left[F(\bm{u}^{g}, \bm{V}^{g+1}) \!\!- \!\!F(\bm{u}^{g},\bm{V}^{g})\right]\!\leq\! -\frac{\eta_{\bm{v}}\tau_{\bm{v}}\!\!\sum\limits_{k=1}^{K}\!\!\left\|\nabla_{\bm{v}}F_{k}(\bm{u}^{g},\bm{v}_{k}^{g})\right\|^2}{8}\nonumber\\
    &+\frac{\eta_{\bm{v}}^{2}\tau_{\bm{v}}^{2}\hat{\sigma}_{\bm{v}}^{2}L_{{\bm{v}}}}{2} + 4\eta_{\bm{v}}^{3}L_{{\bm{v}}}^2\hat{\sigma}_{\bm{v}}^{2}\tau_{\bm{v}}^{2}(\tau_{\bm{v}}-1).
\end{align}
The second line in (64) corresponds to the effect of the $\bm{u}$-step, however, deriving the upper bound of it is more challenging. In particular, the smoothness bound for the $\bm{u}$-step is written as
\begin{align}
    F(\bm{u}^{g+1}\!\!, \!\bm{V}^{e+1}) &\!-\! F(\bm{u}^{g}\!\!, \bm{V}^{g+1})\!\leq\!\langle\nabla_{\bm{u}}F(\bm{u}^{g},\!\bm{V}^{g+1}), \bm{u}^{g+1} \!- \!\bm{u}^{g}\rangle \nonumber\\
    &+ \frac{L_{\bm{u}}}{2}\left\|\bm{u}^{g+1} - \bm{u}^{g}\right\|^{2}.
\end{align}
Several Lemmas are introduced as follows before deriving the smoothness bound in (66).

\textbf{Lemma 2:} Based on Assumption 2 and Assumption 3, for the global part in the $g$th global communication round, we derive that
\begin{align}
    &\sum_{t=1}^{\tau_{\bm{u}}}\sum_{k=1}^{K}\mathbb{E}\|\bm{u}_{k}^{g,t-1} - \bm{u}_{k}^{g}\|^2\nonumber\\
    &\leq \eta_{\bm{u}}^2 \phi_{\bm{u}}^2K\tau_{\bm{u}}^3 + 2\tau_{\bm{u}}D^2\sum_{k=1}^{K}\rho_{k}^{g}.
\end{align}
\begin{proof} In (67), $\bm{u}_{k}^{g}$ is the original global part received from the edge server, and $(\bm{u}_{k}^{g,t-1} - \bm{u}_{k}^{g})$ has two parts, namely, variation obtained from global part updating $(\bm{u}_{k}^{g,t-1} - \bm{u}_{k}^{g,0})$ in the $k$th edge device and variation because of pruning $(\bm{u}_{n}^{g,0} - \bm{u}_{k}^{g})$. Therefore, (67) is rewritten as

\begin{align}
    &\sum_{t=1}^{\tau_{\bm{u}}}\sum_{k=1}^{K}\mathbb{E}\|\bm{u}_{k}^{g,t-1} - \bm{u}_{k}^{g}\|^2 \nonumber\\ 
    & =\sum_{t=1}^{\tau_{\bm{u}}}\sum_{k=1}^{K}\mathbb{E}\|(\bm{u}_{k}^{g,t-1} \!\!-\! \bm{u}_{k}^{g,0}) + (\bm{u}_{k}^{g,0}\!\! - \!\bm{u}_{k}^{g})\|^2 \nonumber\\
    \leq &\sum_{t=1}^{\tau_{\bm{u}}}\sum_{k=1}^{K}2\mathbb{E}\|\bm{u}_{k}^{g,t-1} - \bm{u}_{k}^{g,0}\|^2 + \nonumber\\
    &\sum_{t=1}^{\tau_{\bm{u}}}\sum_{k=1}^{K}2\mathbb{E}\|\bm{u}_{k}^{g,0} - \bm{u}_{k}^{g}\|^2.
\end{align}

In (68), $\bm{u}_{k}^{g,t-1}$ is calculated from $\bm{u}_{k}^{g,0}$ by updating $t-1$ iterations through stochastic gradient descent on the $k$th edge device. Through the local gradient updating, we obtain that

\begin{align}
    &\sum_{t=1}^{\tau_{\bm{u}}}\sum_{k=1}^{K}2\mathbb{E}\|\bm{u}_{k}^{g,t-1} - \bm{u}_{k}^{g,0}\|^2\nonumber\\
    & = 2\sum_{t=1}^{\tau_{\bm{u}}}\sum_{k=1}^{K}\mathbb{E}\left\|\sum_{i=0}^{t-2}-\eta_{\bm{u}}\nabla_{\bm{u}} F_k(\bm{u}_{k}^{g,i}, \bm{v}_{k}^{g,\tau_{\bm{v}}}, \xi_{k}^{g,i})\odot\bm{m}_{k}^{g}\right\|^2\nonumber\\
    &\leq 2\eta_{\bm{u}}^2\sum_{t=1}^{\tau_{\bm{u}}}\sum_{k=1}^{K}(t-1)\sum_{i=0}^{t-2}\mathbb{E}\|\nabla F_k(\bm{u}_{k}^{g,i}, \bm{v}_{k}^{g,\tau_{\bm{v}}},\xi_{k}^{g,i})\odot\bm{m}_{k}^{g}\|^2\nonumber\\
    &\leq 2\eta_{\bm{u}}^2 \phi_{\bm{u}}^2K\sum_{t=1}^{\tau_{\bm{u}}}(t-1)^2
    =\eta^2 \phi_{\bm{u}}^2K\frac{2\tau_{\bm{u}}^3 - 3\tau_{\bm{u}}^2 + \tau_{\bm{u}}}{3}\nonumber\\
    &\leq\eta^2 \phi_{\bm{u}}^2K\tau_{\bm{u}}^3.
\end{align}
In (69), the third step is derived by the upper bound of the second moments of stochastic gradients of the global part in Assumption 3. Then, $\bm{u}_{k}^{g,0} - \bm{u}_{k}^{g}$ in (68) is calculated as
\begin{small}

\begin{align}
    &\sum_{t=1}^{\tau_{\bm{u}}}\sum_{k=1}^{K}2\mathbb{E}\|\bm{u}_{k}^{g,0} - \bm{u}_{k}^{g}\|^2 \nonumber\\
    &=\sum_{t=1}^{\tau_{\bm{u}}}\sum_{k=1}^{K}2\mathbb{E}\|\bm{u}_{k}^{g} \odot\bm{m}_{k}^{g} - \bm{u}_{k}^{g}\|^2 \nonumber\\
    &\leq 2\sum_{t=1}^{\tau_{\bm{u}}}\sum_{k=1}^{K}\rho_{k}^{g}D^2 = 2\tau_{\bm{u}}D^2\sum_{k=1}^{K}\rho_{k}^{g}.
\end{align}
\end{small}
In (70), the second step is derived by the upper bound of the model error caused by the pruning ratio in Assumption 2.
By introducing (69) and (70) into (68), we achieve the upper bound in (67).
\end{proof}

\textbf{Lemma 3:} According to Assumption 1, Assumption 2, and Assumption 3, the difference of the gradient of the global part at the $g$th global communication round is derived as 
\begin{align}
    &\mathbb{E}\left\|\frac{1}{\kappa_{g}^{j}}\sum_{t=1}^{\tau_{\bm{u}}}\sum_{k\in\mathcal{K}_{g}^{j}}[\nabla_{\bm{u}} F_{k}^{j}(\bm{u}_{k}^{g,t-1},\bm{v}_{k}^{g,\tau_{\bm{v}}}) - \nabla_{\bm{u}} F_{k}^{j}(\bm{u}_{k}^{g},\bm{v}_{k}^{g,\tau_{\bm{v}}})]\right\|^2 \nonumber\\
    &\leq \frac{\phi_{\bm{u}}^2K\eta_{\bm{u}}^2L_{\bm{u}}^2\tau_{{\bm{u}}}^4 + 2\tau_{{\bm{u}}}^2L_{\bm{u}}^2D^2\sum_{k=1}^{K}\rho_{k}^{g}}{\kappa^{*}},
\end{align}
where $\kappa_{g}^{j} = |\mathcal{K}_{g}^{j}|$ is the number of the global part keeping the $j$th weight and $\nabla_{\bm{u}} F_{k}^{j}(\bm{u}_{k}^{g},\bm{v}_{k}^{g,\tau_{\bm{v}}})$ is the gradient of the $j$th weight.

\begin{proof}
\begin{small}
\begin{align}
    &\mathbb{E}\left\|\frac{1}{\kappa_{g}^{j}}\sum_{t=1}^{\tau_{\bm{u}}}\sum_{k\in\mathcal{K}_{g}^{j}}[\nabla_{\bm{u}} F_{k}^{j}(\bm{u}_{k}^{g,t-1},\bm{v}_{k}^{g,\tau_{\bm{v}}}) - \nabla_{\bm{u}} F_{k}^{j}(\bm{u}_{k}^{g},\bm{v}_{k}^{g,\tau_{\bm{v}}})]\right\|^2\nonumber\\
    &\leq \frac{\tau_{\bm{u}}}{\kappa_{g}^{j}}\!\!\sum_{t=1}^{\tau_{\bm{u}}}\!\sum_{k\in\mathcal{K}_{g}^{j}}\!\!\mathbb{E}\|\nabla_{\bm{u}} F_{k}^{j}(\bm{u}_{k}^{g,t-1},\bm{v}_{k}^{g,\tau_{\bm{v}}})\! -\! \nabla_{\bm{u}} F_{k}^{j}(\bm{u}_{k}^{g},\bm{v}_{k}^{g,\tau_{\bm{v}}})\|^2\nonumber\\
    &\leq\frac{\tau_{\bm{u}}}{\kappa^{*}}\sum_{t=1}^{\tau_{\bm{u}}}\sum_{k=1}^{K}\mathbb{E}\|\nabla_{\bm{u}} F_{k}^{j}(\bm{u}_{k}^{g,t-1},\bm{v}_{k}^{g,\tau_{\bm{v}}}) - \nabla_{\bm{u}} F_{k}^{j}(\bm{u}_{k}^{g},\bm{v}_{k}^{g,\tau_{\bm{v}}})\|^2\nonumber\\
    &\leq\frac{\tau_{\bm{u}}}{\kappa^{*}}\sum_{t=1}^{\tau_{\bm{u}}}\sum_{k=1}^{K}\mathbb{E}\|\nabla_{\bm{u}} F_{k}(\bm{u}_{k}^{g,t-1},\bm{v}_{k}^{g,\tau_{\bm{v}}}) - \nabla_{\bm{u}} F_{k}(\bm{u}_{k}^{g},\bm{v}_{k}^{g,\tau_{\bm{v}}})\|^2\nonumber\\
    &\leq \frac{\tau_{\bm{u}}}{\kappa^{*}}\sum_{t=1}^{\tau_{\bm{u}}}\sum_{k=1}^{K}L_{\bm{u}}^2\mathbb{E}\|\bm{u}_{k}^{g,t-1} - \bm{u}_{k}^{g}\|^2.
\end{align}
\end{small}
In (72), through relaxing the inequality of the first step by setting $\kappa^{*} = \min\kappa_{g}^{j}$ and assuming the global part of all $K$ edge devices have the $j$th weight, namely, $|\mathcal{K}_{g}^{j}| = K$, the first step is deduced as the second step. Then, due to the fact that the $l_2$-gradient norm of a vector is no larger than the sum of the norm of all sub-vectors, we further relax the sub-vector $\nabla_{\bm{u}}F_{k}^{j}$ of the second step into  $\nabla_{\bm{u}} F_{k}$ and obtain the third step. 
\end{proof}

\textbf{Lemma 4:} Based on the unbiased and bounded variance of the global part in Assumption 4, we derive that 
\begin{small}
\begin{align}
    &\!\!\mathbb{E}\!\left\|\!\frac{1}{\kappa_{g}^{j}}\!\!\sum_{t=1}^{\tau_{\bm{u}}}\!\sum_{k\in\mathcal{K}_{g}^{j}}\!\![\nabla_{\bm{u}} F_{k}^{j}(\bm{u}_{k}^{g,t-1}\!\!\!,\bm{v}_{k}^{g,\tau_{\bm{v}}}\!,\! \xi_{k}^{g,t-1}\!) \!\!- \!\!\nabla_{\bm{u}} F_{k}^{j}(\bm{u}_{k}^{g,t-1}\!,\!\bm{v}_{k}^{g,\tau_{\bm{v}}}\!)]\right\|^2\nonumber\\
    &\leq\frac{\tau_{\bm{u}}^2K\hat{\sigma}_{\bm{u}}^2}{\kappa^{*}}.
\end{align}
\end{small}
\begin{proof}
\begin{small}
\begin{align}
    &\mathbb{E}\left\|\frac{1}{\kappa_{g}^{j}}\sum_{t=1}^{\tau_{\bm{u}}}\sum_{k\in\mathcal{K}_{g}^{j}}[\nabla_{\bm{u}} F_{k}^{j}(\bm{u}_{k}^{g,t-1},\bm{v}_{k}^{g,\tau_{\bm{v}}}, \xi_{k}^{g,t-1}) \right.\nonumber\\
    &- \nabla_{\bm{u}} F_{k}^{j}(\bm{u}_{k}^{g,t-1},\bm{v}_{k}^{g,\tau_{\bm{v}}})]\Bigg\|^2\nonumber\\
    &\leq\frac{\tau_{\bm{u}}}{\kappa^{*}}\sum_{t=1}^{\tau_{\bm{u}}}\sum_{k=1}^K\mathbb{E}\left\|\nabla_{\bm{u}} F_{k}^{j}(\bm{u}_{k}^{g,t-1},\bm{v}_{k}^{g,\tau_{\bm{v}}}, \xi_{k}^{g,t-1}) \right.\nonumber\\
    &- \nabla_{\bm{u}} F_{k}^{j}(\bm{u}_{k}^{g,t-1},\bm{v}_{k}^{g,\tau_{\bm{v}}})\|^2\nonumber\\
    &\leq\frac{\tau_{\bm{u}}}{\kappa^{*}}\sum_{t=1}^{\tau_{\bm{u}}}\sum_{k=1}^K\mathbb{E}\left\|\nabla_{\bm{u}} F_{k}(\bm{u}_{k}^{g,t-1},\bm{v}_{k}^{g,\tau_{\bm{v}}}, \xi_{k}^{g,t-1}) \right.\nonumber\\
    &-\nabla_{\bm{u}} F_{k}(\bm{u}_{k}^{g,t-1},\bm{v}_{k}^{g,\tau_{\bm{v}}})\|^2\nonumber\\
    &\leq\frac{\tau_{\bm{u}}^2K\hat{\sigma}^2}{\kappa^{*}}.
\end{align}
\end{small}
In (74), same as the second step in (72), we relax the sub-vector $\nabla_{\bm{u}}F_{k}^{j}$ of the first step into  $\nabla_{\bm{u}} F_{k}$ and obtain the second step. According to the bounded variance of the global part in Assumption 4, we obtain the upper bound in the last step in (74).
\end{proof}

\textbf{Lemma 5:} The upper bound of the difference of the global part in continuous global communication rounds $\mathbb{E}\|\bm{u}^{g+1}-\bm{u}^{g}\|^2$ is denoted as
\begin{align}
    &\mathbb{E}\|\bm{u}^{g+1}-\bm{u}^{g}\|^2 \leq 3\eta_{\bm{u}}^2N^2\tau_{\bm{u}}^2\phi_{\bm{u}}^2 \nonumber\\
    &+\frac{3N^2\eta_{\bm{u}}^2\tau_{\bm{u}}^2K\hat{\sigma}_{\bm{u}}^2 + 3N^2L_{\bm{u}}^2\tau_{\bm{u}}^4\eta_{\bm{u}}^4K\phi_{\bm{u}}^2}{\kappa^{*}}\nonumber\\
    &+\frac{6N^2\eta_{\bm{u}}^2 L_{\bm{u}}^2D^2\tau_{\bm{u}}^2\sum_{k=1}^{K}\rho_{k}^{g}}{\kappa^{*}}.
\end{align}

\begin{proof}
\begin{small}
\begin{align}
    &\mathbb{E}\|\bm{u}^{g+1}-\bm{u}^{g}\|^2\nonumber\\
    &\!\!=\mathbb{E}\left\|\sum_{j=1}^{N}\frac{1}{\kappa_{g}^{j}}\sum_{k\in\mathcal{K}_{g}^{j}}\sum_{t=1}^{\tau_{\bm{u}}}\eta_{\bm{u}}\nabla_{\bm{u}} F_{k}^{j}(\bm{u}_{k}^{g,t-1},\bm{v}_{k}^{g,\tau_{\bm{v}}}, \xi_{k}^{g,t-1})\right\|^2\nonumber\\
    &\leq 3N\sum_{j=1}^{N}\!\mathbb{E}\!\left\|\frac{1}{\kappa_{g}^{j}}\!\!\sum_{k\in\mathcal{K}_{g}^{j}}\sum_{t=1}^{\tau_{\bm{u}}}\!\!\eta_{\bm{u}}[\nabla_{\bm{u}} F_{k}^{j}(\bm{u}_{k}^{g,t-1},\bm{v}_{k}^{g,\tau_{\bm{v}}}\!, \xi_{k}^{g,t-1})\!\! \right.\nonumber\\
    &- \nabla_{\bm{u}} F_{k}^{j}(\bm{u}_{k}^{g,t-1},\bm{v}_{k}^{g,\tau_{\bm{v}}})]\Bigg\|^2\nonumber\\
    &+3N\sum_{j=1}^{N}\mathbb{E}\left\|\frac{1}{\kappa_{g}^{j}}\sum_{k\in\mathcal{K}_{g}^{j}}\sum_{t=1}^{\tau_{\bm{u}}}\eta_{\bm{u}}[\nabla_{\bm{u}} F_{k}^{j}(\bm{u}_{k}^{g,t-1},\bm{v}_{k}^{g,\tau_{\bm{v}}}) \right.\nonumber\\
    &- \nabla_{\bm{u}} F_{k}^{j}(\bm{u}_{k}^{g},\bm{v}_{k}^{g,\tau_{\bm{v}}})]\Bigg\|^2\nonumber\\
    &+3N\sum_{j=1}^{N}\mathbb{E}\left\|\frac{1}{\kappa_{g}^{j}}\sum_{k\in\mathcal{K}_{g}^{j}}\sum_{t=1}^{\tau_{\bm{u}}}\eta_{\bm{u}}[\nabla_{\bm{u}} F_{k}^{j}(\bm{u}_{k}^{g},\bm{v}_{k}^{g,\tau_{\bm{v}}})]\right\|^2,
\end{align}
\end{small}
where we split stochastic gradient $\nabla_{\bm{u}} F_{k}^{j}(\bm{u}_{k}^{g,t-1},\bm{v}_{k}^{g,\tau_{\bm{v}}}, \xi_{k}^{g,t-1})$ into three parts, namely, $[\nabla_{\bm{u}} F_{k}^{j}(\bm{u}_{k}^{g,t-1},\bm{v}_{k}^{g,\tau_{\bm{v}}}, \xi_{k}^{g,t-1}) - \nabla_{\bm{u}} F_{k}^{j}(\bm{u}_{k}^{g,t-1},\bm{v}_{k}^{g,\tau_{\bm{v}}})]$, $[\nabla_{\bm{u}} F_{k}^{j}(\bm{u}_{k}^{g,t-1},\bm{v}_{k}^{g,\tau_{\bm{v}}}) - \nabla_{\bm{u}} F_{k}^{j}(\bm{u}_{k}^{g},\bm{v}_{k}^{g,\tau_{\bm{v}}})]$, and $[\nabla_{\bm{u}} F_{k}^{j}(\bm{u}_{k}^{g},\bm{v}_{k}^{g,\tau_{\bm{v}}})]$.

According to the upper bound of the second moments of stochastic gradients of the global part, the upper bound of the third term in (76) is calculated as
\begin{align}
    &3N\sum_{j=1}^{N}\mathbb{E}\left\|\frac{1}{\kappa_{g}^{j}}\sum_{k\in\mathcal{K}_{g}^{j}}\sum_{t=1}^{\tau_{\bm{u}}}\eta_{\bm{u}}[\nabla_{\bm{u}} F_{k}^{j}(\bm{u}_{k}^{g},\bm{v}_{k}^{g,\tau_{\bm{v}}})]\right\|^2\nonumber\\
    &\leq 3\eta_{\bm{u}}^2 N\tau_{\bm{u}}\sum_{j=1}^{N}\sum_{t=1}^{\tau_{\bm{u}}}\mathbb{E}\|\nabla_{\bm{u}} F_{k}(\bm{u}_{k}^{g},\bm{v}_{k}^{g,\tau_{\bm{v}}})\|^2\nonumber\\
    &\leq 3\eta_{\bm{u}}^2N^2\tau_{\bm{u}}^2\phi_{\bm{u}}^2.
\end{align}
Through plugging (71), (73), and (77) into (76), the upper bound of $\mathbb{E}\|\bm{u}^{g+1}-\bm{u}^{g}\|^2$ is derived as (75).
\end{proof}
Based on the aforementioned Lemmas 2-5 of the global part, we further derive the upper bound of the smoothness bound for the $\bm{u}$-step in (64). We first consider taking expectations on both sides of (64) and achieve
\begin{align}
    &\mathbb{E}[F(\bm{u}^{g+1}, \bm{V}^{g+1})] - \mathbb{E}[F(\bm{u}^{g}, \bm{V}^{g+1})] \nonumber\\
    &\leq\mathbb{E}\langle\nabla_{\bm{u}}F(\bm{u}^{g},\!\bm{V}^{g+1}), \bm{u}^{g+1} \!- \!\bm{u}^{g}\rangle + \frac{
    L_{\bm{u}}}{2}\mathbb{E}\|\bm{u}^{g+1}-\bm{u}^{g}\|^2.
\end{align}
In (78), $\mathbb{E}\langle\nabla_{\bm{u}}F(\bm{u}^{g},\!\bm{V}^{g+1}), \bm{u}^{g+1} \!- \!\bm{u}^{g}\rangle$ is equal to a sum of inner products over all model weights of the global part and is written as
\begin{align}
    &\mathbb{E}\langle\nabla_{\bm{u}}F(\bm{u}^{g},\!\bm{V}^{g+1}), \bm{u}^{g+1} - \bm{u}^{g}\rangle \nonumber\\
    &= \sum_{j=1}^{N}\mathbb{E}\langle\nabla F^{j}(\bm{u}^{g},\bm{V}^{g+1}), \bm{u}^{g+1,j}-\bm{u}^{g,j}\rangle\nonumber\\
    &=\sum_{j=1}^{N}\mathbb{E}\Bigg\langle\nabla_{\bm{u}} F^{j}(\bm{u}^{g},\bm{V}^{g+1}),\nonumber\\&-\frac{1}{\kappa_{g}^{j}}\sum_{k\in\mathcal{K}_{g}^{j}}\sum_{t=1}^{\tau_{\bm{u}}}\eta_{\bm{u}}\nabla_{\bm{u}} F_{k}^{j}(\bm{u}_{k}^{g,t-1},\bm{v}_{k}^{g,\tau_{\bm{v}}})\Bigg\rangle\nonumber\\
    &= - \sum_{j=1}^{N}\mathbb{E}\langle\nabla_{\bm{u}} F^{j}(\bm{u}^{g},\bm{V}^{g+1}), \eta_{\bm{u}} \tau_{\bm{u}}\nabla_{\bm{u}} F^{j}(\bm{u}^{g},\bm{V}^{g+1})\rangle\nonumber\\
    &- \sum_{j=1}^{N}\mathbb{E}\Bigg\langle\nabla_{\bm{u}} F^{j}(\bm{u}^{g},\bm{V}^{g+1}),\nonumber\\
    &\frac{1}{\kappa_{g}^{j}}\!\!\sum_{k\in\mathcal{K}_{g}^{j}}\sum_{t=1}^{\tau_{\bm{u}}}\!\eta_{\bm{u}}\left[\nabla_{\bm{u}} F_{k}^{j}(\bm{u}_{k}^{g,t-1},\bm{v}_{k}^{g,\tau_{\bm{v}}}) - \nabla_{\bm{u}} F_{k}^{j}(\bm{u}_{k}^{g},\bm{v}_{k}^{g,\tau_{\bm{v}}})\right]\!\!\Bigg\rangle.
\end{align}
In (79), considering a reference point $\eta_{\bm{u}} \tau_{\bm{u}}\nabla_{\bm{u}} F^{j}(\bm{u}^{g},\bm{V}^{g+1})$ of the personalized part, the result of the last step is divided into two parts, and according to the Peter-Paul inequality in (61), the first term in (79) of the last step is further computed as
\begin{align}
    &- \sum_{j=1}^{N}\mathbb{E}\langle\eta_{\bm{u}} \tau_{\bm{u}}\nabla_{\bm{u}} F^{j}(\bm{u}^{g},\bm{V}^{g+1}), \eta_{\bm{u}} \tau_{\bm{u}}\nabla_{\bm{u}} F^{j}(\bm{u}^{g},\bm{V}^{g+1})\rangle \nonumber\\
    &= -\eta_{\bm{u}} \tau_{\bm{u}}\sum_{j=1}^{N}\left\|\nabla_{\bm{u}} F^{j}(\bm{u}^{g},\bm{V}^{g+1})\right\|^2.
\end{align}
Then, according to the Peter-Paul inequality in (62) with constant $m=1$ and Lemma 3, the second term in (79) of the last step is calculated as
\begin{align}
    &- \sum_{j=1}^{N}\mathbb{E}\Bigg\langle\nabla F^{j}(\bm{u}^{g},\bm{V}^{g+1}),\nonumber\\
    &\left.\frac{1}{\kappa_{g}^{j}}\sum_{k\in\mathcal{K}_{g}^{j}}\!\sum_{t=1}^{\tau_{\bm{u}}}\eta_{\bm{u}}\left[\nabla_{\bm{u}} F_{k}^{j}(\bm{u}_{k}^{g,t-1},\bm{v}_{k}^{g,\tau_{\bm{v}}})\!\! -\!\! \nabla_{\bm{u}} F_{k}^{j}(\bm{u}_{k}^{g},\bm{v}_{k}^{g,\tau_{\bm{v}}})\right]\!\right\rangle\nonumber\\
    &= - \sum_{j=1}^{N}\eta_{\bm{u}} \tau_{\bm{u}}\mathbb{E}\Bigg\langle\nabla_{\bm{u}} F^{j}(\bm{u}^{g},\bm{V}^{g+1}),\nonumber\\
    &\left.\frac{1}{\tau_{\bm{u}}}\frac{1}{\kappa_{g}^{j}}\sum_{k\in\mathcal{K}_{g}^{j}}\!\sum_{t=1}^{\tau_{\bm{u}}}\left[\nabla_{\bm{u}} F_{k}^{j}(\bm{u}_{k}^{g,t-1},\bm{v}_{k}^{g,\tau_{\bm{v}}})\!\! -\!\! \nabla_{\bm{u}} F_{k}^{j}(\bm{u}_{k}^{g},\bm{v}_{k}^{g,\tau_{\bm{v}}})\right]\!\!\right\rangle\nonumber\\
    &\leq\frac{\eta_{\bm{u}}\tau_{\bm{u}} }{2}\sum_{j=1}^{N}\mathbb{E}\|\nabla_{\bm{u}} F^{j}(\bm{u}^{g},\bm{V}^{g+1})\|^2 + \nonumber\\
    &\frac{\eta_{\bm{u}}}{2\tau_{\bm{u}}}\sum_{j=1}^{N}\mathbb{E}\left\|\frac{1}{\kappa_{g}^{j}}\sum_{k\in\mathcal{K}_{g}^{j}}\sum_{t=1}^{\tau_{\bm{u}}}[\nabla_{\bm{u}} F_{k}^{j}(\bm{u}_{k}^{g,t-1},\bm{v}_{k}^{g,\tau_{\bm{v}}})\right.\nonumber\\
    &-\!\! \nabla_{\bm{u}} F_{k}^{j}(\bm{u}_{k}^{g},\bm{v}_{k}^{g,\tau_{\bm{v}}})]\Bigg\|^2.\nonumber\\
    &\leq\frac{\eta_{\bm{u}}\tau_{\bm{u}}}{2}\sum_{j=1}^{N}\mathbb{E}\|\nabla_{\bm{u}} F^{j}(\bm{u}^{g},\bm{V}^{g+1})\|^2 \nonumber\\
    &+\frac{N\phi_{\bm{u}}^2K\eta_{\bm{u}}^3L_{\bm{u}}^2\tau_{\bm{u}}^3 + 2N\eta_{\bm{u}} \tau_{\bm{u}}L_{\bm{u}}^2D^2\sum_{k=1}^{K}\rho_{k}^{g}}{2\kappa^{*}}.
\end{align}
By plugging (80) and (81) into (79), $\mathbb{E}\langle\nabla_{\bm{u}}F(\bm{u}^{g},\!\bm{V}^{g+1}), \bm{u}^{g+1} - \bm{u}^{g}\rangle$ is derived as
\begin{align}
    &\mathbb{E}\langle\nabla_{\bm{u}}F(\bm{u}^{g},\bm{V}^{g+1}), \bm{u}^{g+1} - \bm{u}^{g}\rangle\leq \nonumber\\
    &-\frac{\eta_{\bm{u}}\tau_{\bm{u}}}{2}\sum_{j=1}^{N}\left\|\nabla_{\bm{u}} F^{j}(\bm{u}^{g},\!\bm{V}^{g+1})\right\|^2\nonumber\\
    &+\frac{N\phi_{\bm{u}}^2K\eta_{\bm{u}}^3L_{\bm{u}}^2\tau_{\bm{u}}^3 + 2N\eta_{\bm{u}} \tau_{\bm{u}}L_{\bm{u}}^2D^2\sum_{k=1}^{K}\rho_{k}^{g}}{2\kappa^{*}}.
\end{align}
Finally, based on the derived upper bound of  $\mathbb{E}\|\bm{u}^{g+1} - \bm{u}^{g}\|^2$ and (82), the upper bound of the smoothness bound for the $\bm{u}$-step in (78) is calculated as
\begin{align}
    &\mathbb{E}[F(\bm{u}^{g+1}, \bm{V}^{g+1})] - \mathbb{E}[F(\bm{u}^{g}, \bm{V}^{g+1})] \nonumber\\
    &\leq -\frac{\eta_{\bm{u}}\tau_{\bm{u}}}{2}\sum_{j=1}^{N}\left\|\nabla_{\bm{u}} F^{j}(\bm{u}^{g},\!\bm{V}^{g+1})\right\|^2 + \frac{L_{\bm{u}}}{2}\mathbb{E}\|\bm{u}^{g+1} - \bm{u}^{g}\|^2\nonumber\\
    &+\frac{N\phi_{\bm{u}}^2K\eta_{\bm{u}}^3L_{\bm{u}}^2\tau_{\bm{u}}^3 + 2N\eta_{\bm{u}} \tau_{\bm{u}}L_{\bm{u}}^2D^2\sum_{k=1}^{K}\rho_{k}^{g}}{2\kappa^{*}}.
\end{align}
By taking expectations on both sides of (64) and plugging (65) and (82) into it, we achieve
\begin{align}
    &\mathbb{E}[F(\bm{u}^{g+1}, \bm{V}^{g+1}) - F(\bm{u}^{g},\bm{V}^{g})]\nonumber\\
    &\leq-\frac{\eta_{\bm{v}}\tau_{\bm{v}}\!\!\sum\limits_{k=1}^{K}\!\!\left\|\nabla_{\bm{v}}F_{k}(\bm{u}^{g},\bm{v}_{k}^{g})\right\|^2}{8}\nonumber\\
    &+\frac{\eta_{\bm{v}}^{2}\tau_{\bm{v}}^{2}\hat{\sigma}_{\bm{v}}^{2}L_{{\bm{v}}}}{2} + 4\eta_{\bm{v}}^{3}L_{{\bm{v}}}^2\hat{\sigma}_{\bm{v}}^{2}\tau_{\bm{v}}^{2}(\tau_{\bm{v}}-1)\nonumber\\
    & -\frac{\eta_{\bm{u}}\tau_{\bm{u}}}{2}\sum_{j=1}^{N}\left\|\nabla_{\bm{u}} F^{j}(\bm{u}^{g},\!\bm{V}^{g+1})\right\|^2 \nonumber\\
    &+\frac{N\phi_{\bm{u}}^2K\eta_{\bm{u}}^3L_{\bm{u}}^2\tau_{\bm{u}}^3 + 2N\eta_{\bm{u}} \tau_{\bm{u}}L_{\bm{u}}^2D^2\sum_{k=1}^{K}\rho_{k}^{g}}{2\kappa^{*}}\nonumber\\
    &+\frac{3\eta_{\bm{u}}^2N^2\tau_{\bm{u}}^2\phi_{\bm{u}}^2L_{\bm{u}}}{2} +\frac{3N^2\eta_{\bm{u}}^2\tau_{\bm{u}}^2K\hat{\sigma}_{\bm{u}}^2L_{\bm{u}} + 3N^2L_{\bm{u}}^3\tau_{\bm{u}}^4\eta_{\bm{u}}^4K\phi_{\bm{u}}^2}{2\kappa^{*}}\nonumber\\
    &+\frac{3N^2\eta_{\bm{u}}^2 L_{\bm{u}}^3D^2\tau_{\bm{u}}^2\sum_{k=1}^{K}\rho_{k}^{g}}{\kappa^{*}}.
\end{align}
Furthermore, by considering all $G$ global communication rounds in (84), we achieve
\begin{align}
    &\frac{1}{G}\sum_{g=1}^{G}\left[\frac{\eta_{\bm{v}}\tau_{\bm{v}}}{8}\sum\limits_{k=1}^{K}\|\nabla_{\bm{v}}F_{k}(\bm{u}^{g},\bm{v}_{k}^{g})\|^2 \right.\nonumber\\
    &+ \frac{\eta_{\bm{u}}\tau_{\bm{u}}}{2}\sum_{j=1}^{N}\|\nabla_{\bm{u}} F^{j}(\bm{u}^{g},\!\bm{V}^{g+1})\|^2\Bigg]\nonumber\\
    &\leq \frac{\mathbb{E}[F(\bm{u}^{0}, \bm{V}^{0}) - F(\bm{u}^{*},\bm{V}^{*})]}{G}+\frac{\eta_{\bm{v}}^{2}\tau_{\bm{v}}^{2}\hat{\sigma}_{\bm{v}}^{2}L_{{\bm{v}}}}{2} \nonumber\\
    &+ 4\eta_{\bm{v}}^{3}L_{{\bm{v}}}^2\hat{\sigma}_{\bm{v}}^{2}\tau_{\bm{v}}^{2}(\tau_{\bm{v}}-1)+\frac{3\eta_{\bm{u}}^2N^2\tau_{\bm{u}}^2\phi_{\bm{u}}^2L_{\bm{u}}}{2} \nonumber\\
    &+ \!\frac{N\phi_{\bm{u}}^2K\eta_{\bm{u}}^3L_{\bm{u}}^2\tau_{\bm{u}}^3\!+\!3N^2\eta_{\bm{u}}^2\tau_{\bm{u}}^2K\hat{\sigma}_{\bm{u}}^2L_{\bm{u}} \!+ \!3N^2L_{\bm{u}}^3\tau_{\bm{u}}^4\eta_{\bm{u}}^4K\phi_{\bm{u}}^2}{2\kappa^{*}}\nonumber\\
    &+\frac{(N\eta_{\bm{u}} \tau_{\bm{u}}L_{\bm{u}}^2D^2+3N^2\eta_{\bm{u}}^2 L_{\bm{u}}^3D^2\tau_{\bm{u}}^2)\sum_{g=1}^{G}\sum_{k=1}^{K}\rho_{k}^{g}}{G\kappa^{*}}.
\end{align}
%





\ifCLASSOPTIONcaptionsoff
  \newpage
\fi





\bibliographystyle{IEEEtran}
\bibliography{IEEEabrv,Ref,ReferencesMP}
%
%

\end{document}